\documentclass{article} 
\usepackage{iclr2026_conference,times}

\DeclareUnicodeCharacter{2581}{\rule{.5em}{.4pt}}
\usepackage{titletoc}

\usepackage{amsmath,amsfonts,bm}

\def\eqref#1{equation~\ref{#1}}

\def\1{\bm{1}}

\DeclareMathAlphabet{\mathsfit}{\encodingdefault}{\sfdefault}{m}{sl}
\SetMathAlphabet{\mathsfit}{bold}{\encodingdefault}{\sfdefault}{bx}{n}

\usepackage{spverbatim}
\usepackage{graphicx}
\usepackage{booktabs}
\usepackage{algorithm}
\usepackage{tabularx}
\usepackage{wrapfig}
\usepackage{amsmath}
\usepackage{amssymb} 
\usepackage{algpseudocode}
\usepackage{multirow}

\usepackage{hyperref}
\usepackage{url}

\usepackage{xcolor}
\usepackage{xfrac}

\newcommand{\llava}{LLaVA}
\newcommand{\llama}{LLaMA}
\newcommand{\vidllama}{VideoLLaMA3}
\newcommand{\llavavid}{LLaVA-Video}

\title{Linear Mechanisms for Spatiotemporal Reasoning in Vision Language Models}

\author{Raphi Kang$^*$, Hongqiao Chen{}\thanks{denotes equal contribution}{  
 }, Georgia Gkioxari, Pietro Perona \\
California Institute of Technology\\
\texttt{\{rkang, harrychen, georgia, perona\}@caltech.edu} 
}

\iclrfinalcopy 

\begin{document}

\maketitle

\begin{abstract}

Spatio-temporal reasoning is a remarkable capability of Vision Language Models (VLMs), but the underlying mechanisms of such abilities remain largely opaque. 
We postulate that visual/geometrical and textual representations of spatial structure must be combined at some point in VLM computations. We search for such confluence, and ask whether the identified representation can causally explain aspects of input-output model behavior through a linear model.
We show empirically that VLMs encode object locations by linearly binding \textit{spatial IDs} to textual activations, then perform reasoning via language tokens.
Through rigorous causal interventions we demonstrate that these IDs, which are ubiquitous across the model, can systematically mediate model beliefs at intermediate VLM layers.
Additionally, we find that spatial IDs serve as a diagnostic tool for identifying limitations in existing VLMs, and as a valuable learning signal.
We extend our analysis to video VLMs and identify an analogous linear temporal ID mechanism.
By characterizing our proposed spatiotemporal ID mechanism, we elucidate a previously underexplored internal reasoning process in VLMs, toward improved interpretability and the principled design of more aligned and capable models. We release our code for reproducibility: \url{https://github.com/Raphoo/linear-mech-vlms}.

\end{abstract}

\vspace{-2mm}
\begin{figure}[h]
    \centering
\includegraphics[width=0.9\linewidth]{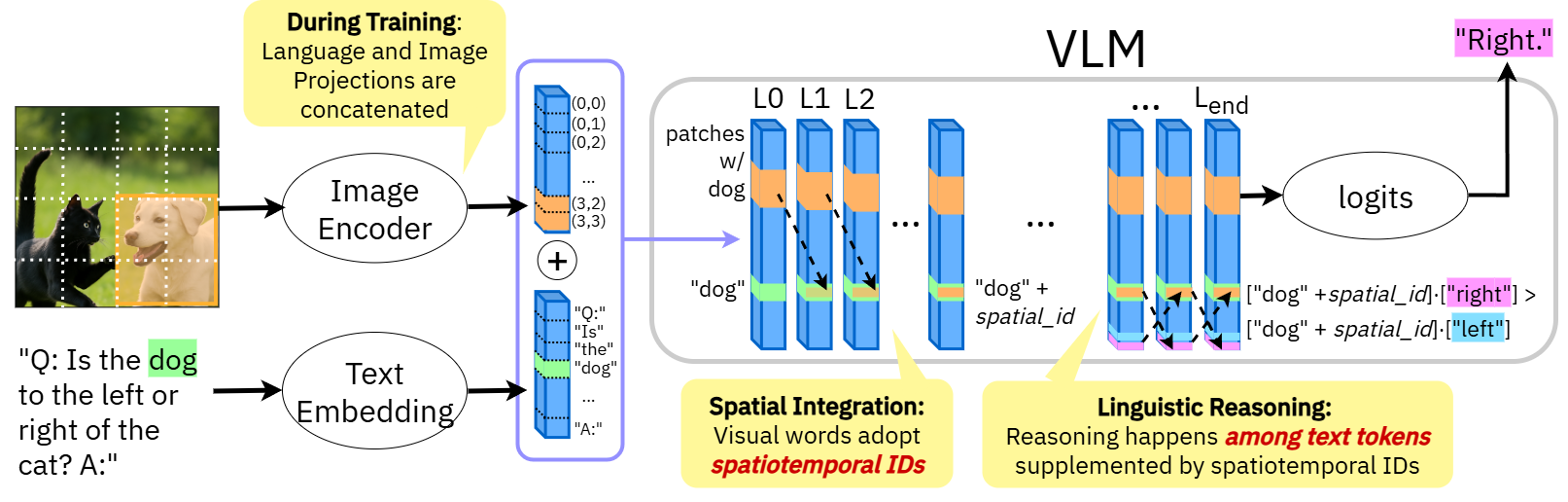}
\vspace{-2mm}
    \caption{{\bf Hypothesis for spatiotemporal visual reasoning}. The VLM linearly binds spatiotemporal localization to object word activations in early layers. Subsequent linguistic reasoning about the object is informed by its location in space and time per the spatiotemporal ID. }
    \vspace{-6mm}
    \label{fig:main}
\end{figure}

\section{Introduction}
\label{Intro}

Reasoning about visual input with textual queries is a key challenge behind vision-language models (VLMs). Consider a typical visual question answering (VQA) prompt: “\textit{Is the dog to the left or right of the cat?}”. To succeed at this, one must resolve linguistic references, locate entities in the visual field, assess spatial relationships, and make a categorical decision. 
Though complex capabilities in spatial or temporal reasoning are still far from being fully understood or reliably engineered \citep{stogiannidis2025mind, chen2025spatial, tong2024eyes}, SoTA VLMs have seen steady progress in simple visual reasoning without explicit guidance. So how do they do it?

Attention-based analyses in VLMs have shown various structural properties emerge in VLM internals during VQA \citep{jiang_devils_2025, neo_towards_2024, zhang_redundancy_2024}. Relatedly, mechanistic interpretability in LLMs has uncovered linear circuits for relational linguistic reasoning \citep{park_linear_2024, feng_how_2024, hernandez_linearity_2024}. Might such linear processes also be driving visual reasoning in VLMs, and if so, how exactly?
This leads us to ask:  
\textbf{Q1.} \textit{Can we linearly model emergent structured reasoning processes that drive spatial reasoning in VLM internals?}

\begin{figure}
    \centering\includegraphics[width=0.98\linewidth]{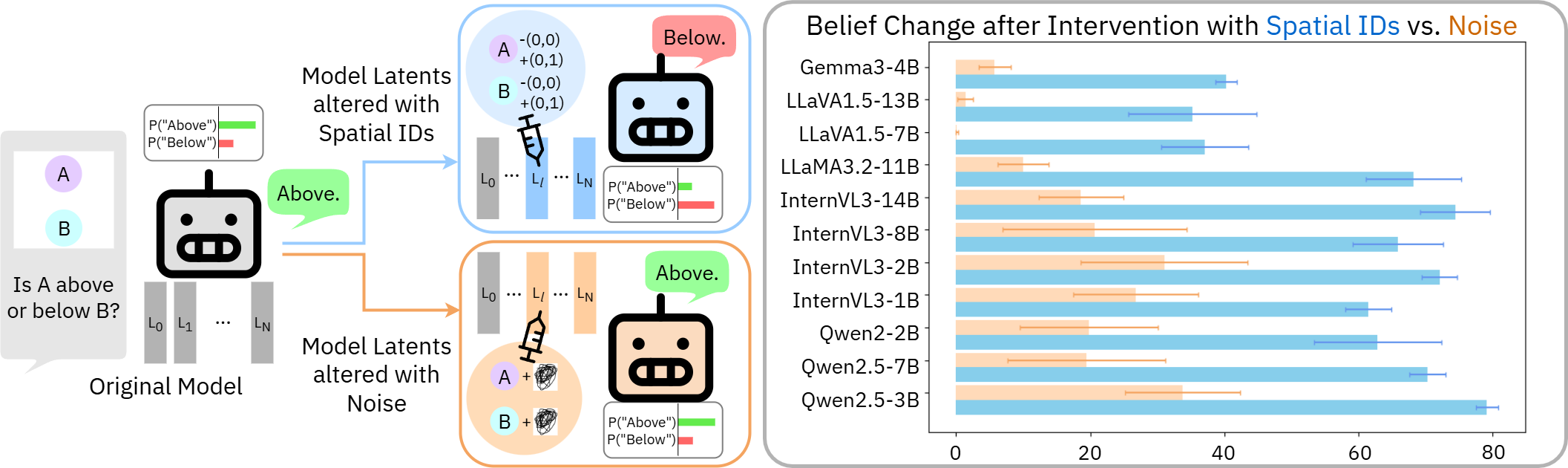}\vspace{-4mm}
    \caption{
    {\bf Results from Targeted Intervention} (\S \ref{sec:causality}). Median binary belief swap due to spatial ID steering is 64.4\%, and 29.5\% for noise. Spatial IDs have 43.6\% above-chance influence on average.
    We conclude that spatial IDs mediate models' beliefs about objects' locations in space.
    }
    \label{fig:all_models}
    \vspace{-4mm}
\end{figure}



The typical VLM architecture utilizes a vision encoder which projects the input image to embeddings that are prepended to  text token embeddings. This is then processed by a downstream vision-aligned LLM.
Popular model families using this paradigm are \llava \citep{liu_visual_2023}, \llama \citep{dubey2024llama}, Qwen \citep{bai_qwen25-vl_2025}, InternVL \citep{chen_internvl_2024}, and Gemma \citep{team2024gemma}. A growing body of work aims to improve spatial reasoning capacities in VLMs \citep{chen2024spatialvlm, fan_grit_2025} and temporal reasoning in video models \citep{xiao2024can, li2024temporal}. 
Identification of the internal mechanism by which SoTA VLMs do spatial VQA can empower engineers to identify current architectural components leading to VQA failure modes in 3D reasoning or simple VQA.
To this end, we ask:
\textbf{Q2.} \textit{Given our linear model of spatial reasoning in model activations, how do we use it to understand and improve SoTA VLMs?}

Similar training paradigms to image-based VLMs yield video models such as \llavavid \citep{zhang_video_2024}, \vidllama \citep{zhang_videollama_2025}, and  Qwen2.5 \citep{bai_qwen25-vl_2025}, among others. Given our theory for the mechanisms underlying spatial reasoning in VLMs, we ask: \textbf{Q3.} \textit{Do video models  utilize analogous linear mechanisms for temporal reasoning?}

To address these questions, we conduct a mechanistic analysis of autoregressive VLMs and construct a linear model for spatiotemporal reasoning. 
We show that VLMs decompose a visual reasoning task by first binding spatial information about visual objects to object word activations, 
in the form of linear components we term \textit{spatial IDs}, answering Q1 (Fig. \ref{fig:main}). 
We then extract these IDs and demonstrate their mediative capacity on model output through targeted belief steering in text activations (Fig. \ref{fig:all_models}). We further find that spatial IDs provide insight on VLMs' struggle with depth reasoning, and incorrect spatial IDs as a result of weak vision encoder or poor modality integration leads to failures in LLaVA and LLaMA. This answers Q2. Finally, we show that temporal IDs similarly mediate video models, answering Q3. In summary, our novel contributions are:



\vspace{-2mm}
\begin{itemize}
    \item \textbf{Spatial ID Model Formulation}: We propose a linear model of spatial reasoning in VLMs, called \textit{spatial IDs}. These are text-anchored latent structures that bind visual elements to object tokens thus enabling linguistic reasoning about space (\S \ref{sec information flow}). We emprically extract them from SoTA VLMs for characterization (\S \ref{sec: empirical_extraction}).
\vspace{-1mm}
    \item \textbf{Analytical and Empirical Proof of Causality}: We show model belief can be manipulated by perturbing only the spatial IDs, demonstrating their causal role in reasoning (\S \ref{sec:causality}), and provide theoretical intuition for the emergence of spatial IDs in VLMs (\S \ref{sec theoretical sketch}).
\vspace{-1mm}
    \item \textbf{SoTA VLM Analysis with Spatial IDs}: Through targeted intervention, we identify limitations in depth expression (\S \ref{sec depth rep}) and systematic failure modes in LLaMA/LLaVA (\S \ref{sec diagnosis}), and find models can be effectively finetuned with spatial ID guidance (\S \ref{improving VLMs}.).
\vspace{-1mm}
    \item \textbf{Extension to Temporal IDs in Video Models}: We perform our extraction and characterization analysis on SoTA video models and show that linear temporal IDs, like spatial IDs, can drive temporal reasoning in VLMs (\S \ref{sec video temporal IDs}).
\end{itemize}
\vspace{-3mm}

\section{Emergent structure in Spatial Visual Reasoning}
\label{sec: derive_spatial_ids}

In this section, we characterize the spatial reasoning circuits in SoTA VLMs and isolate any linearly separable components used to communicate spatial information. Towards this end, we track information flow in VLMs and identify important junctions for spatial information transfer across token sequences. Then we empirically extract linear spatial IDs, and analytically derive how they arise.

\subsection{Tracking information flow during reasoning}
\label{sec information flow}

To uncover whether VLMs engage in structured visual reasoning, i.e., isolating and propagating spatial information across layers, we intervene on internal activations during inference. 


\begin{figure}

    \centering
    \includegraphics[width=0.9\linewidth]{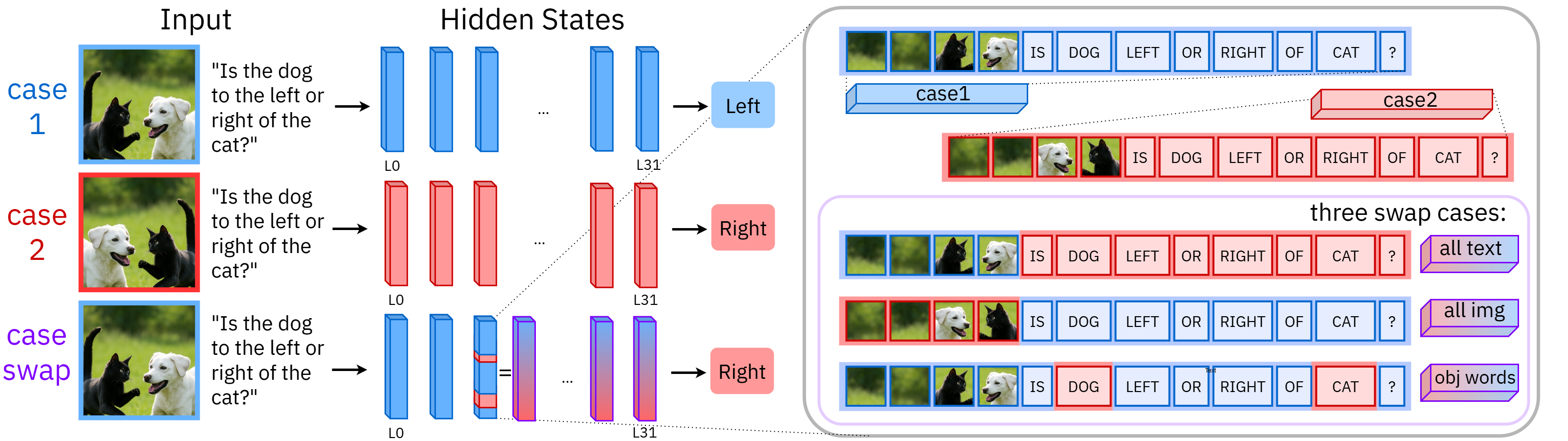}
    \vspace{-4mm}
    \caption{{\bf Mirror swapping experiment} (\S \ref{sec information flow}). Activations from case 1 and 2 are partially swapped at a select layer, in one of three arrangements. 
    Computations continue normally after this point.}
    \vspace{-3mm}
    \label{fig:swapping-illust}
\end{figure}

\textbf{Mirror Swapping Experiment}. 
Our goal is to compare the model’s output when presented with two distinct images and the same text query.  If the model uses localized intermediate representations to reason about spatial relationships, then swapping activations between spatially distinct inputs at key layers and sequence indices should disrupt its final belief about spatial orientation, while swaps between spatially equal but attribute-wise different inputs shouldn't have a strong effect.

Concretely, we run inference on plain and mirrored image-text pairs, extract their representations $x$ at an intermediate layer $L$, then replace a subset $Q$ of activations in the original $x_L$ with activations from the mirrored counterpart $y_L$. The modified representation $\tilde{x}_L$ is passed through the remaining layers. We conduct interventions with three variants of $Q$: (1) all text tokens (2) all image patches (3) object-word tokens only. 
If information critical to spatial reasoning is concentrated in any of these, the model’s belief will change when that region is overwritten.
As a control, we concurrently perform ``attribute swapping", which follows the same steps but instead of mirroring the input image for the intervention case, changes its colors. 
  The intervention procedure is visualized in Fig. \ref{fig:swapping-illust} and formally defined in Alg. \ref{alg:intervene_swaps_mirror}. Further implementation specifics are deferred to Appendix \S \ref{appendix:mirror swapping section}.
\begin{algorithm}[h]
\caption{Swapping Intermediate Activations in Mirrored Images}
\label{alg:intervene_swaps_mirror}
\begin{algorithmic}
\State $x_L , y_L\gets f_L \circ \cdots \circ f_1(x)$ ,\quad $f_L \circ \cdots \circ f_1(y)$ \Comment{x,y: [seq\_dim, embed\_dim]}
\State $\tilde{x}_L \gets x_L[\tilde Q] +  y_L[Q]  $ \Comment{$\tilde{x}_L$: [seq\_dim, embed\_dim], Q: [num\_of\_inds]}
\State $\tilde{x}_{\text{out},L}, {y}_{\text{out}} \gets f_{L_{\max}} \circ \cdots \circ f_{L+1}(\tilde{x}_L)$
 , \quad$ f_{L_{\max}} \circ \cdots \circ f_{L+1}(y_L)$ \Comment{$P_{\tilde{x}_{\text{out},L}}(\text{``GT"})$: [1]}
\end{algorithmic}
\end{algorithm}
\vspace{-3mm}

Here, $Q$ denotes the selected indices in the input sequence to swap, and $\tilde Q$ is all other indices. 
We use the \textsc{COCO-Spatial} benchmark \citep{kamath2023s} for the mirrored images, which is a curated subset of COCO \citep{lin2014microsoft} annotated with spatial language.
To quantify belief shift caused by the intervention, we compute the fraction of the mirror-induced change that can be attributed to the swapped activations at layer $L$.
For the ground truth logit ``GT", this quantity is derived as:
\begin{equation}
   \text{belief shift}_L = \scriptscriptstyle{\frac{P_{x_{\text{out}}}(\text{``GT"}) - P_{\tilde{x}_{\text{out}, L}}(\text{``GT"})}{P_{x_{\text{out}}}(\text{``GT"}) - P_{y_{\text{out}}}(\text{``GT"})}}
\end{equation}

\begin{figure}
    \centering
    \includegraphics[width=1.00\linewidth]{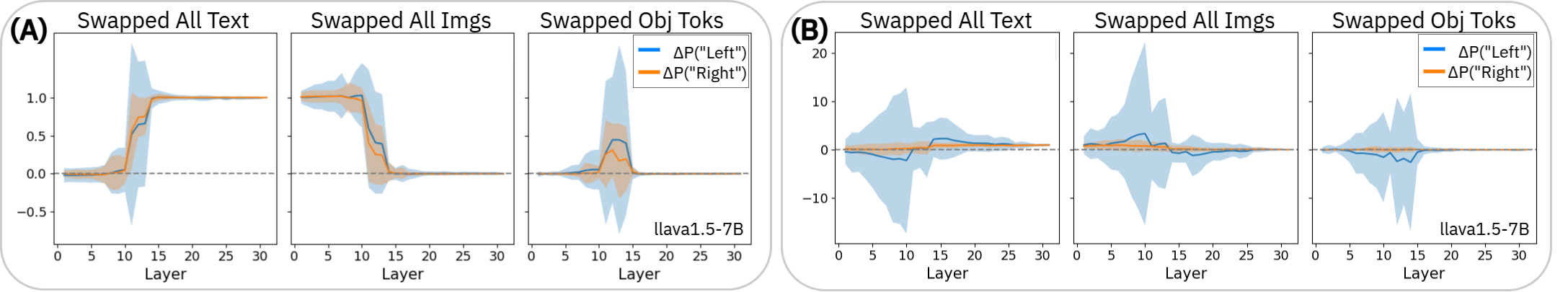}
    \vspace{-8mm}
    \caption{\textbf{Ratio change in log probability for logits ``left" and ``right" from mirror swap }(A) \textbf{and attribute swap} (B) interventions. (A) shows distinct binary belief swaps, where text tokens have an influence after middle layers. Image patches stop having an influence after that point, and object word tokens \textit{only} have an influence in these middle layers. The control, (B), is noisy.}
\vspace{-4mm}
    \label{fig:Steering-res}
\end{figure}

\textbf{Results from Mirror Swapping} are shown in Fig. \ref{fig:Steering-res}A.
Through mirror swapping, we observe a \textit{layer-specific effect} for intervention effect across modalities. Intervening on visual patch tokens has a strong effect in early layers but fades with depth. Conversely, interventions on text tokens increasingly affect final outputs in later layers. This is corroborated by observations that middle layers have a modality switching effect in VLMs \citep{jiang_devils_2025}.  
Notably, swapping only the object-word tokens alters spatial belief specifically within a narrow band of intermediate layers.

Attribute swapping results (Fig. \ref{fig:Steering-res}B) indicate that mirror swapping is a strong experimental setup for assessing spatial information flow in isolation from spurious visual factors.
For the belief shift metric, a value of 0.0 on the y axis indicates model belief in the intervened case is equivalent to case 1 (original query), while 1.0 indicates the belief is equivalent to case 2 (mirrored/changed query).  Mirror swapping results in distinct and strong binary belief swaps whereas attribute swapping yields mostly noise, to the point belief shift magnitudes are -20$\sim$20x that of the original belief difference.


These results suggest that VLMs extract and encode spatial facts from the image into object word tokens' activations, then operate over them in text-space. 
We term the latent structures holding visual spatial information \textit{spatial ids}.  
Inspired by latest mechanistic interpretability findings (discussed in \S \ref{related_work}), we hypothesize that the manner of spatial information storage is approximately linear.



\subsection{Empirical Derivation of Spatial IDs}
\label{sec: empirical_extraction}

 
If spatial IDs are indeed linearly bound to object word activations, we should be able to extract them by averaging out object-related semantics from text activations. Below we outline the process of their extraction. In \S \ref{sec:causality}, we will test if these IDs causally mediate model beliefs, to validate whether the spatial reasoning mechanism in VLMs is indeed linear. 
\vspace{-4mm}
\begin{wrapfigure}{r}{0.45\textwidth}
  \begin{center}
\includegraphics[width=0.45\textwidth]{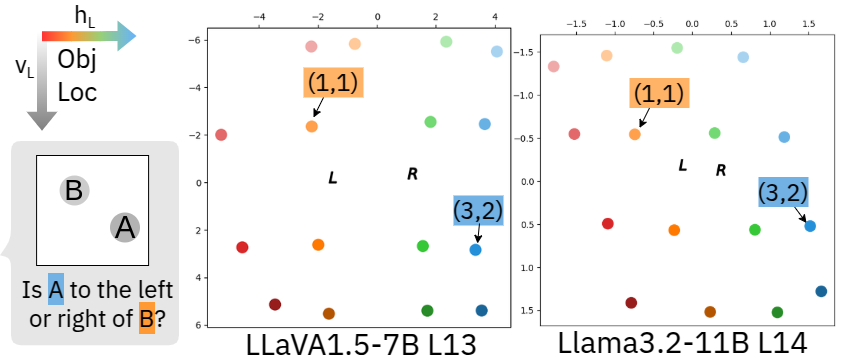}
  \end{center}
  \vspace{-5mm}
    \caption{\textbf{Spatial IDs in a grid.} 
    Color and saturation of markers represent the location of the object when spatial ID was extracted.
     x and y axes are coefficients of ID projections onto $h_L$ and $v_L$. L, R represent ``left", ``right" textual activations.}
    \label{fig:spatial_grid}
    \vspace{-8mm}
\end{wrapfigure}





\textbf{Extraction Preliminaries}.  We first set up some formalisms to derive spatial IDs. 
Let \( \mathcal{O} = \{o_1, o_2, \dots, o_N\} \) denote a set of object categories. For each object \( o \in \mathcal{O} \), we have a set of images \( \{I_{(i,j)}\} \) where the object is positioned at spatial coordinates \( (i,j) \) in a \( m \times m \) grid. Then let \( T^{(o)} \) be a natural language query containing the token corresponding to object \( o \), such as ``Is there an \(o\)?''. We define 
\( \phi_L(o; I_{(i,j)}^{(o)}, T^{(o)}) \in \mathbb{R}^d \) as the embedding of the text token corresponding to object \( o \), extracted from layer \( L \) of the VLM when input= \( (I_{(i,j)}^{(o)}, T^{(o)}) \).
The mean embedding for object \( o \) at layer \( L \) is then:
\vspace{-3mm}
\begin{equation}
        \bar{\phi}_L^{(o)} = \frac{1}{m^2} \sum_{i=0}^{m-1} \sum_{j=0}^{m-1} \phi_L(o; I_{(i,j)}^{(o)}, T^{(o)})
\label{equation: first}
\end{equation}
\vspace{-3mm}

Yielding the object-specific spatial ID at location \( (i,j) \) for object \( o \):
\vspace{-1mm}
\begin{equation}
\label{eq:object-specific-extract}
        \Delta_L^{(o)}(i,j) = \phi_L(o; I_{(i,j)}^{(o)}, T^{(o)}) - \bar{\phi}_L^{(o)}
\end{equation}
\vspace{-5mm}

From this we can derive the \textit{universal spatial ID} at location \( (i,j) \), averaged over N objects.
\vspace{-2mm}
\begin{equation}
        \Delta_L(i,j) = \frac{1}{N} \sum_{n=1}^{N} \Delta_L^{(o_n)}(i,j)
\end{equation}
\label{eq:general extract}
\vspace{-6mm}

To extract canonical horizontal and vertical directions from the universal spatial IDs \( \Delta_L(i,j) \in \mathbb{R}^d \), we compute average difference vectors across grid-aligned coordinate pairs. The vertical and horizontal direction vectors \( v_L, h_L  \in \mathbb{R}^d \), corresponding to increasing \( i \) and \(j \), are computed based on the spatial IDs. Eq. \ref{equation:direction vectors} shows the derivation for $v_L$, and $h_L$ is derived in an analogous manner.
\begin{equation}
\begin{aligned}
    v_L = \frac{1}{m \cdot \binom{m}{2} }\sum_{i=0}^{m-1} \sum_{j_1 > j_2} \left[ \Delta_L(i, j_1) - \Delta_L(i, j_2) \right]
\label{equation:direction vectors}
\end{aligned}
\end{equation}




\vspace{-3mm}
\textbf{Empirical Extraction}. For our study, we extract spatial IDs from 11 SoTA VLMs, with synthetic images created from open-source \textsc{Objaverse} \citep{deitke2023objaverse} objects. The object renders are paired onto a grid of $m = 4$ on top of random natural backgrounds. We provide further extraction details in Appendix \S\ref{appendix sec: spatial ID extract instruct}, along with ablations showing extracted spatial IDs are invariant to chosen images \S\ref{supp sec ablations} and counterfactual studies confirming that spatial IDs reside in object words, and spatial axes are orthogonal \S\ref{supp sec counterfactuals}.
Fig. \ref{fig:spatial_grid} shows two example spatial ID grids projected onto their respective spatial vectors. IDs from more models are shown in \S\ref{exp res on more models appendix sec}. We see that these extracted IDs arrange in an approximate $m \times m$ grid at modality binding layers.
Also projected are activations for spatial words, where we find that ``left" is closer to leftmost spatial IDs, and ``right" vice versa.



\subsection{Theoretical Sketch of Spatial IDs}
\label{sec theoretical sketch}
We now offer a quick, highly minimal analytical intuition for how the emergence of spatial IDs can be ubiquitous across many different models. 
Let
$p=(i,j)$ be some coordinate on a $m\times m$ grid.
Then for some query to a VLM, let the input sequence contain
projected visual tokens $\{x_p\} \text{ for all } p$, and the query text tokens include an object
token $o$.
The residual update to $o$ by one head is:
\begin{equation}
    r_o \leftarrow r_o + W_{\text{out}} \sum_{p \in P}\alpha_{o\leftarrow p}\,v_p,
    \qquad 
    \alpha_{o\leftarrow p} \propto \exp\!\Big(\tfrac{q_o^\top k_p}{\sqrt d}\Big), 
    \quad v_p = W_V x_p .
\end{equation}
\vspace{-5mm}

With cross-modal alignment, attention peaks at the true object patch $p^\star$, giving 
$\delta r_o \approx W_{\text{out}} W_V x_{p^\star}$.
Decompose each patch as $
    x_p = s_p + P\,\psi(p) + \varepsilon_p $,
where $s_p$ encodes content, 
$\psi(p) \in \mathbb{R}^{d_\psi}$ is a shared positional basis (e.g.\ RoPE or learned 2D embeddings), 
$P$ maps positional features into model space, 
and $\varepsilon_p$ is small.
We can now substitute $\phi_L(o;I_{p^\star}, T^{(o)}) = r_{o} + \delta r_{o,p^\star}$ into Eq. \ref{eq:object-specific-extract}. A detailed derivation is in \S\ref{sec: empirical_extraction}, but in summary we get:
\begin{equation}\label{eq:spatial-id}
    \Delta_L(p^\star) = \Delta_L(i,j) \;\approx\; 
    \underbrace{W_{\text{out}} W_V P}_{M \text{ (fixed per model)}}\,
    \Big(\psi(i,j) - \tfrac{1}{m^2}\sum_{p}\psi(p)\Big).
\end{equation}
\vspace{-5mm}

 Thus, spatial IDs are approximately a linear transformation of a universal positional basis written into the object token by attention. 
 Spatial logits are thus approximately linear readouts:
\begin{equation}
    \ell(\textsc{left}) - \ell(\textsc{right}) 
    \;\approx\; (w_{\textsc{left}} - w_{\textsc{right}})^\top \Delta_L(i,j),
\end{equation}
so if $(w_{\textsc{left}} - w_{\textsc{right}})^\top M$ aligns with the $x$-coordinate in $\psi$, the model prefers ``left.'' 
Empirically, a low-rank linear fit from positional encodings $\psi$ to spatial IDs $\Delta_L$ explains most variance (e.g.\ rank-3 gives $R^2 \gtrsim 0.85$, see \S \ref{appendix sec empirical_spatial_id_posenc}, Table~1). 
A more detailed derivation for $ \Delta_L(i,j)$ for the multihead case is shown in Appendix \S \ref{supp sec informal_proof}.
This is a particularly simplified setting, and real reasoning circuits in VLMs will involve a lot more noise and nonlinearities.
The main takeaway is that VLM designs like Fig. \ref{fig:main} encourage models to endow text tokens with visual information, followed by lingustic reasoning. This information transfer, in its most simplified linear form, is in the form of spatial IDs. 

In practice, the finegrained circuit employed by VLMs may be much more varied, distributed, and nonlinear. The spatial ID framework could capture just one component of a more complex system. But per Ockham's Razor, spatial IDs are powerful due to their simplicity. In following sections, we demonstrate the mediative influence of this simple spatial ID model on final VLM outputs, and further show how spatial IDs can be leveraged to improve existing models and build stronger ones.

\section{Spatial IDs Mediate Model Beliefs}
\label{sec:causality}


If spatial IDs capture the causal mechanisms behind spatial reasoning, we should be able to linearly subtract or add arbitrary IDs to object word activations and change the model's belief about object location. In this section, we design and perform experiments on real naturalistic images to test that empirically derived spatiotemporal IDs have causal effects on model outputs on spatial VQA.

\textbf{Steering with Arbitrary IDs Experiment}. For some layer $L$, we denote the model residuals corresponding to the entire input sequence after that layer as $x_L$, and perturb its token activation at some index $q$ to observe any effects on the output belief. Alg. \ref{alg:intervene_on_obj} illustrates the process.

\begin{algorithm}
\caption{Intervention at Layer $L$ via Residual Modification}
\label{alg:intervene_on_obj}
\begin{algorithmic}
\State  $x_L \gets f_L \circ \cdots \circ f_1(x)$ \Comment{x: [ seq\_dim, embed\_dim]} 
\State  $\tilde{x}_L \gets x_L[:q] + [x_L[q] + \Delta_{L}(i,j) - \tilde \Delta_{L}(i,j)] + x_L[q+1:]  $ \Comment{$\Delta_{L}(i,j)$: [embed\_dim]}
\State  $\tilde{x}_{\text{out}} \gets f_{L_{\max}} \circ \cdots \circ f_{L+1}(\tilde{x}_L)$ \Comment{$P_{\tilde{x}_{out}}(\text{``GT"})$: [1]}
\end{algorithmic}
\end{algorithm}

Here we scale the norm of $\Delta_{L}(i,j)$ to be $\alpha| x_L[q]|$, and $\tilde \Delta_{L}(i,j) = \Delta_{L}(m-i-1,j)$. This approximately preserves the norm of $x_L$. $\alpha=5$ is some scaling constant set after grid searching for stable intervention.
We take 100 COCO images where one object is to the left or right of another, per labels from \textsc{Coco-Spatial}, and ask queries of the form ``Is x to the left/right of y"?. We measure the log probability of ``left" and ``right" tokens in the final output logits to assess steering effects.


\begin{figure}
    \centering
    \includegraphics[width=0.99\linewidth]{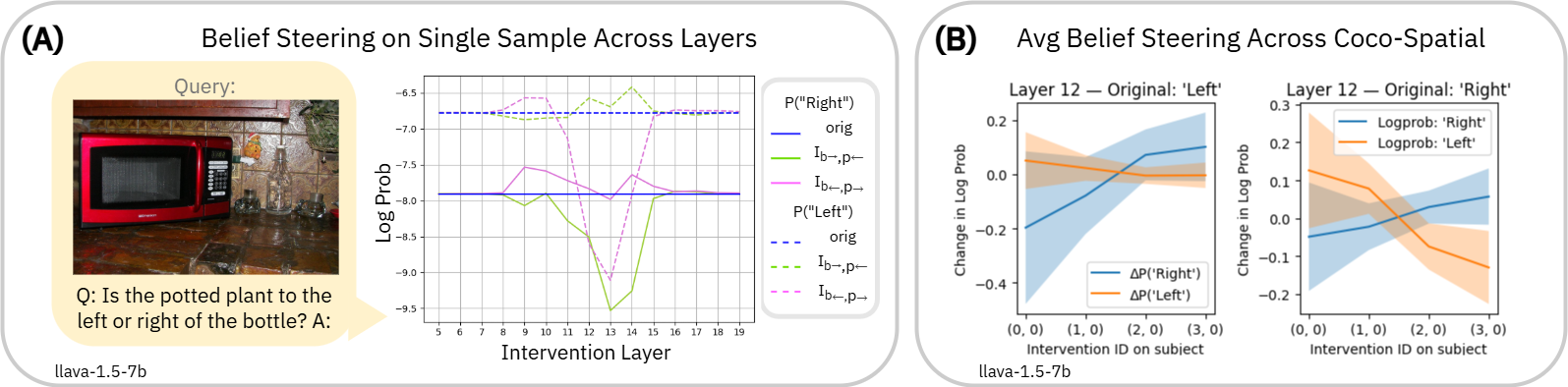}
    \includegraphics[width=0.99\linewidth]{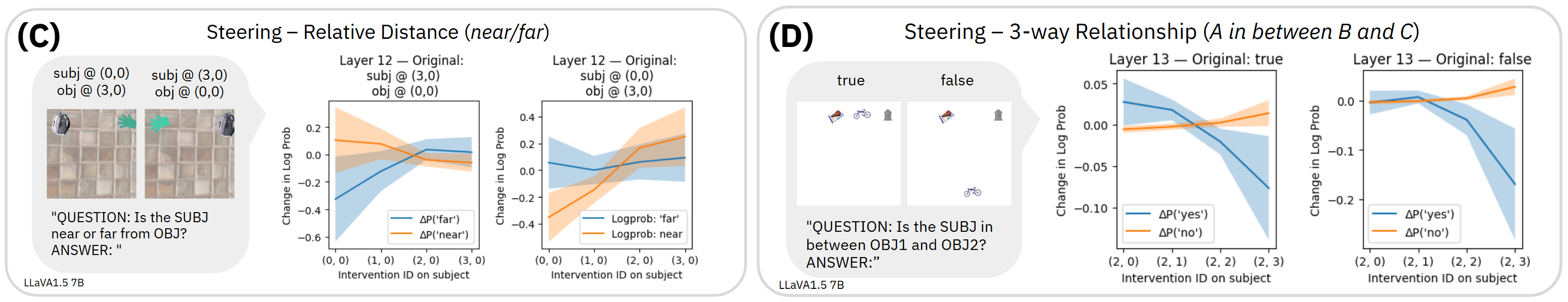}
    \vspace{-4mm}
    \caption{ \textbf{Effect of spatial steering on real images} on one sample across different intervention layers (A) and across a dataset for one layer (B). In (A), dotted and solid lines indicate answer probabilities for ``left" or ``right". Different colors indicate no intervention (blue), steering the bottle to the left and plant to the right (pink), and the reverse steering (green). Blue lines are flat and show that the unintervened model incorrectly assigns a higher log probability to ``left".
Pink lines show intervention on intermediate layers results in overwriting initial incorrect beliefs. 
    (B) shows the shift in log probability for ```left" vs. ``right" as  a result of spatial steering on the subject word token.
    (C) and (D) show shifts in log probability for ``near" vs. ``far",  and ``yes" to an object being sandwiched between two others, vs. ``no".}
    \vspace{-4mm}
    \label{fig:Steering}
\end{figure}

\textbf{Results from Arbitrary Steering}. Fig. \ref{fig:Steering} shows the effects of model belief steering on real images and videos. Fig. \ref{fig:Steering}A shows that steering impact is greatest at modality alignment layers as expected per the mirror swapping analysis, and Fig \ref{fig:Steering}B shows that intervening with the rightmost spatial ID largely enhances model belief that the object is to the right, and vice versa for the leftmost ID for leftward belief.
The y axes show changes in log probability for those binary directions for the whole dataset, and x axes show the different ID locations. Regardless of whether the answer to the original query was ``left" or ``right", subplot trends are the same.

We repeat the analysis for queries about relative distance and three-way relationships where one object is sandwiched \textit{in between} two others. Again, we find that when the object is to the left, altering the spatial ID of the subject towards the right increases the likelihood of ``far" and decreases that of ``near", and vice versa if the object is to the right. Similarly, we find that bringing a subject closer and closer to be surrounded by two objects increases the model's belief that the subject is \textit{in between} the objects.


\textbf{Adversarial Steering Experiment}.  If spatial IDs are indeed ubiquitous across models, interventions on internal activations should change the resultant model beliefs across many SoTA models. 
To confirm this, we evaluate the log probability of the correct answer (``GT") and its opposite (``$\neg$GT") for all samples in \textsc{COCO-spatial} on 11 SoTA models. Then, we repeat this measurement after intervention with spatial IDs most likely to reverse their original beliefs.
More detailed experimental procedure is provided in \S \ref{appendix sec: adversarial steering fig 2}. In addition to targeted adversarial steering, we perform steering with noise vectors of the same norm as the spatial IDs, to evaluate chance belief swaps.

\textbf{Adversarial Steering Results}.
We report \% binary belief swaps on \textsc{COCO-spatial} from the spatial ID vs. noise steering case in Fig. \ref{fig:all_models}. Steering with spatial IDs yields a median 64.6\% swap in beliefs, versus 29.5\% with noise. This indicates activation intervention has nonzero chance influence on model output, but there is a clear above-chance average of 43.6\% increase with spatial IDs. Here, a model's belief on one sample is considered ``swapped" if the relative likelihood of the ground truth and its opposite answer has changed.
For example, if $P(\text{``left"}) > P(\text{``right"})$ before intervention, but after intervention we see $P(\text{``left"}) < P(\text{``right"})$, the intervention has swapped the model belief.
Thus we conclude that spatial ID mechanisms mediate model belief in the models considered.

\section{Spatial IDs for Understanding and Improving Image VLMs}

With the existence and causal nature of spatial IDs established, we explore two ways to leverage them towards stronger VLMs. 
First, we aim to understand why 3D reasoning fails in SoTA VLMs. Second, we use spatial IDs to diagnose architectural bottlenecks of SoTA VLMs in VQA.




\subsection{Depth Representation in Image VLMs}
\label{sec depth rep}

\begin{wrapfigure}{r}{0.65\textwidth}
  \begin{center}
  \vspace{-4mm}
\includegraphics[width=0.65\textwidth]{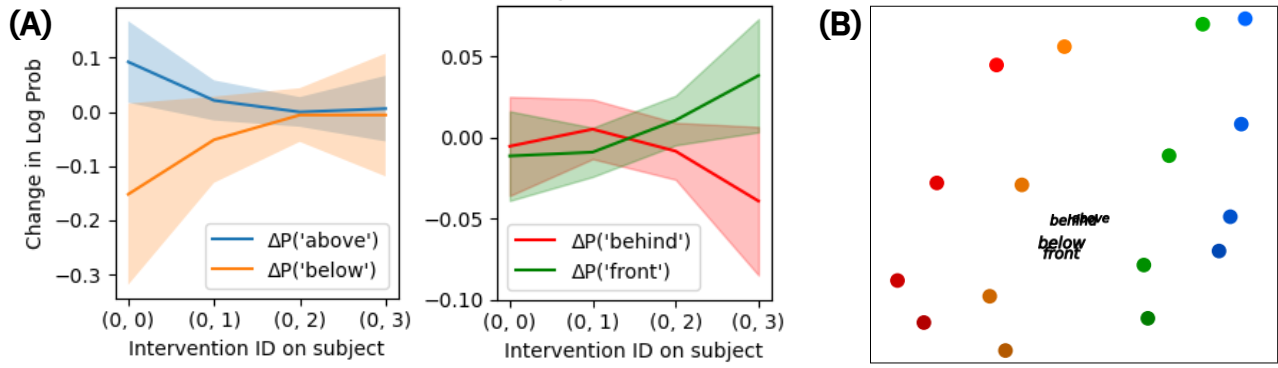}
  \end{center}
  \vspace{-5mm}
  \caption{\textbf{Depth and height are strongly correlated in LLaVA}. (A) Steering results for IDs varying in y-dim and their impact on beliefs about height or depth. (B) Projection of spatial words onto a spatial ID grid. Embeddings for ``above"/ ``front" and``behind"/``below" are nearly identical. }
  \label{fig:placeholder-for-depth}
    \vspace{-3mm}
\end{wrapfigure}

Spatial IDs suggest that VLMs represent visual space within a 2D grid. What might this mean for depth? 
We hypothesize that the language model must reason about depth related queries using the 2D localization in context. To verify whether this is the case, we look at the resulting belief changes in the depth axis when the \llava1.5 7B model is steered with spatial IDs varying in height. 
Fig. \ref{fig:placeholder-for-depth} shows the results. The same spatial IDs increasing the likelihood for ``above" and decreasing ``below", also drive up ``front" and drive down ``behind" in \llava. 

Further, projection of these  word embeddings onto spatial vectors reveals that ``above"/``behind" 
and ``below"/``front" map 
to overlapping locations, indicating their functional relationships with spatial IDs are similar. These results may be due to biases in training, or innate shortcomings in the VLM architecture. They certainly highlight the need for better depth-handling mechanisms, whether that be through improved training data or tooling.

\subsection{Diagnosing VLMs}
\label{sec diagnosis}

When a
VLM fails at a spatial task, how do we pinpoint the reason it failed? Referring back to Fig. \ref{fig:main}, VLM failure points can roughly be divided into modality encoding, crossmodal information integration, or linguistic reasoning stages.
Knowing what part of a VLM's architecture must be improved to reduce failures is paramount to efficient model engineering. 


Per-sample analysis of spatial IDs provides a unique ability to identify a model's bottleneck. 
Consider an evaluation set $\mathbf{K} = \{k_1, k_2 ... k_K\} $, where each $k = (image, query)$.
An imperfect VLM will fail at some samples.
In this section, we perform two experiments to identify the architectural component which causes for the distribution of $\mathbf{K}_{wrong}$ to be statistically distinct from $\mathbf{K}_{correct}$.

An example diagnosis process may look like this. If a model exhibits \textit{incorrect} spatial ID binding, and that incorrect output produced is faithful with the spatial ID, then the language-only reasoning stage is likely not at fault. From there, if a model 
exhibits sensitivity to masking the correct object region
for $\mathbf{K}_{wrong}$ but not for $\mathbf{K}_{correct}$, the vision encoder is the likely bottleneck. If there is no distinct sensitivity difference, the errors are likely taking place after the vision encoder, but before the linguistic reasoning.  If model accuracy seems independent of both spatial ID correctness and image recognition capacity, the language model layers beyond spatial ID binding are likely the biggest bottleneck. Note that it is possible for incorrect spatial IDs to be correlated to wrong answers, but still have some model inaccuracies be resultant from factors other than spatial IDs, such as erroneous priors during LM readout \citep{leng2024mitigating, ramakrishnan2018overcoming}. In this case, it is still valuable to find if models can benefit from stronger spatial representations through this diagnosis process, and minimize avenues for failure.
For the described analyses, we need a sufficient $\mathbf{K}_{wrong}$ subset. As their $\mathbf{K}_{wrong}$ are biggest on \textsc{coco-spatial}, we select LLaVA1.5 7B and LLaMA3.2VL 11B as model organisms for this section.






\textbf{Ground Truth Spatial ID Deviation Experiment}.  First, we want to identify if models predict incorrect spatial IDs for the samples they get wrong. If the answer is \textit{yes}, then it is likely that the downstream language model is not the performance bottleneck, since it is faithful to the spatial information received.
To compute the deviation of the model's believed spatial ID to the ground truth (g.t.), we compute the g.t. spatial ID by 
projecting the word activation onto the spatial axes:
\vspace{-1mm}
\begin{equation}
\label{eq:extracting object}
     \Delta_L^{(o)}(i,j)_{ext} \approx V V^T\phi_L(o; I_{(i,j)}^{(o)}, T^{(o)}), \quad V = [v_L, h_L]
\end{equation}\vspace{-5mm}

For a spatial query like ``Is the $o$ to the left or right of a $\tilde{o}$?", we can thus compute $\Delta^{(o)}(i,j)_{\text{gt}}$ and $\Delta^{(\tilde{o})}(i,j)_{\text{gt}}$. 
The model's assigned spatial IDs to the objects are computed per Eq. \ref{eq:extracting object}, for $\Delta^{(o)}(i,j)_{\text{ext}}$ and $\Delta^{(\tilde{o})}(i,j)_{\text{ext}}$.
Then the g.t. ID margin deviation for some object $o$ is:
\begin{equation}
    \text{ID deviation margin} = \epsilon_{\text{ext}} - \epsilon_{\text{gt}} ,  \quad \text{where  } \epsilon_{\text{gt}} = i^{(o)}_{\text{gt}} - i^{(\tilde{o})}_{\text{gt}}, \epsilon_{\text{ext}} = i^{(o)}_{\text{ext}} - i^{(\tilde{o})}_{\text{ext}}
\end{equation}
\vspace{-4mm}

Here, a negative margin indicates that the model's extracted spatial IDs oppose the ground truth.

\begin{figure}
    \centering
    \includegraphics[width=0.75\linewidth]{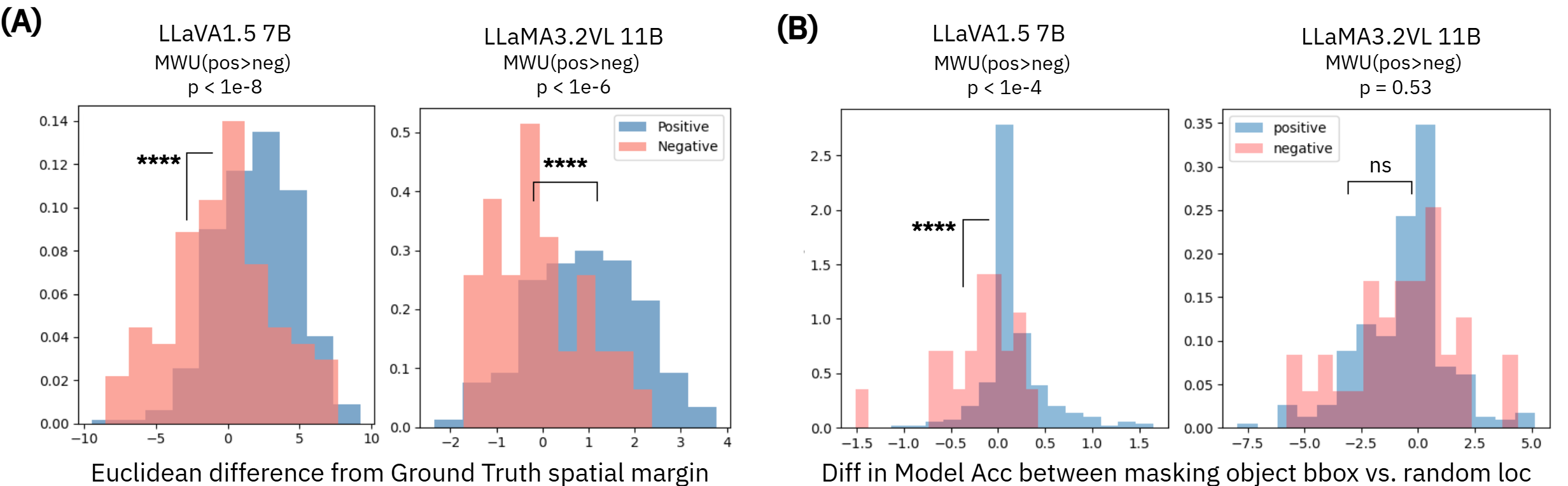}\vspace{-4mm}
    \caption{\textbf{Contrasting density histograms show incorrect spatial IDs drive bad predictions}.  (A) shows deviation of model spatial IDs from g.t., and (B) the difference in model accuracy when masking objects vs. random locations in images. Histograms are samples VLMs got right (blue) or wrong (red). LLaVA shows faulty object detection with wrong answers, while LLaMA does not.
   }
   \vspace{-4mm}
    \label{fig:Diagnosis}
\end{figure}

\textbf{ID Deviation Results}. From Fig. \ref{fig:Diagnosis}A, we see that deviation from ground truth in extracted spatial ID margin is highly correlated with model mistakes. In other words, for LLaVA and LLaMA, wrong spatial IDs in object word activations led to wrong model answers, so linguistic reasoning was not the reason these failures occurred.
Each subplot shows two density histograms overlaid in the same grid, where the x axis is $\epsilon_{\text{ext}} - \epsilon_{\text{gt}}$. The red histogram represents the density of ID deviations for $\mathbf{K}_{wrong}$, and the blue histogram shows the same for $\mathbf{K}_{correct}$. The red distribution is visibly skewed to the negatives compared to the blue. Quantitatively, we perform the Mann-Whitney U test \citep{mcknight2010mann} to calculate the p-value for the hypothesis that the two distributions (red and blue) are non-identical.
 Now we ask, is this failure mode stemming from the vision encoder level, or does it occur during the spatial ID binding across modalities?

\textbf{Image Masking Experiment}.
Altering the raw image input at the pixel level can help us understand whether it is a faulty vision encoder or faulty crossmodal information integration that has led to the failures. If the model's beliefs on $\mathbf{K}_{correct}$ are more sensitive to masking the image raw input at the g.t. location of $o$, while beliefs on $\mathbf{K}_{wrong}$ change more with masking elsewhere, we can conclude that the vision encoder of this VLM is doing a poor job at object detection, leading to observed failures. If we do not observe this is the case, the failure may arise from the crossmodal information integration stage. In other words, the language model is doing a poor job appending binding IDs, despite the vision encoder having the necessary object recognition capacity.


We design an obfuscation paradigm inspired by methods like D-RISE \citep{petsiuk2021black}, where we either blur the bounding box of $o$, or $R$ other locations in the image that do not intersect with the bboxes for $o$ or $\tilde{o}$. We then measure model belief change when masking the object vs. elsewhere:

\vspace{-5mm}
\begin{equation}
    \text{bbox sensitivity} = \big( P(\text{``GT"}) - P(\text{``GT"}|\text{mask }o) \big) - \big( P(\text{``GT"}) - min_r[P(\text{``GT"}|\text{mask }r) \text{, } r \in R] \big) 
\end{equation}
\vspace{-5mm}

\textbf{Image Masking Results}. 
Fig. \ref{fig:Diagnosis}B shows overlaid histograms for bounding box masking sensitivities of $\textbf{K}_{correct}$ and $\textbf{K}_{wrong}$. Here, a negative value indicates greater sensitivity to raw pixel masking of random scenes, suggesting poor object detection. For \llava, there is a statistically significant p-value for the hypothesis that $\textbf{K}_{wrong}$ is shifted more negative than $\textbf{K}_{correct}$, indicating its vision encoder fails at object detection when it answers incorrectly. In contrast, $\mathbf{K}_{wrong}, \mathbf{K}_{correct}$ in LLaMA are agnostic to image obfuscation.
This suggests that its failure modes likely stem after the vision encoder.
These insights could be attributed to how LLaVA uses an out-of-the-box ViT that was text-aligned at a massive scale, hence not being tuned for finegrained detection, while LLaMA has a trained in-house ViT whose image-text alignment may be less robust.

\textbf{Diagnosis Conclusion}. With spatial IDs, we explore the causes for failure in a few model VLMs. We find that for both LLaMA and LLaVA, the linguistic reasoning stage is faithful to spatial IDs. LLaVA's vision encoder is likely creating wrong spatial IDs from poor object detection, while LLaMA's weak point appears to be information integration across the image patch activations to the text tokens.
These conclusions are preliminary and do not suggest that \textit{all} of a model's failures stem from \textit{one} architectural component, but can serve to guide finetuning stage choices when resources are scarce, or provide intuition for future model designs.
\vspace{-3mm}

\begin{wrapfigure}{r}{0.3\textwidth}
  \begin{center}
  \vspace{-3mm}
\includegraphics[width=0.3\textwidth]{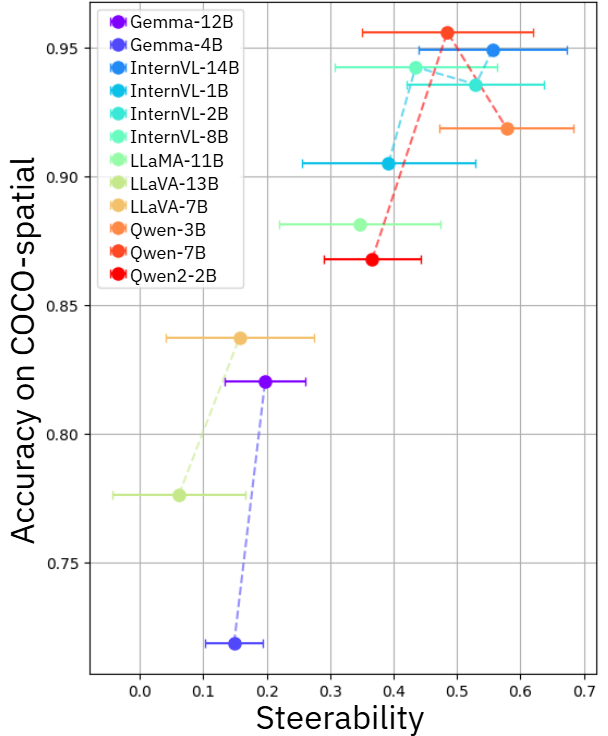}
  \end{center}
  \vspace{-3mm}
\caption{\textbf{Accuracy vs. Steerability}. Models with higher accuracy can be better steered with spatial IDs. }
\label{fig:spatial ID and model performance correlation}
\vspace{-9mm}
\end{wrapfigure}

\subsection{Improving VLMs}
\label{improving VLMs}
\textbf{Spatial IDs and Model Performance}. To understand if spatial IDs could be a valuable learning signal, we first evaluate whether stronger steerability from spatial IDs is correlated to stronger models. Fig. \ref{fig:spatial ID and model performance correlation} shows
the results of this analysis, where indeed we see that models with higher zero-shot accuracy on COCO-spatial also exhibit greater belief changes with spatial ID interventions.

We define “steerability” as the difference between the change of belief resultant from steering with opposing spatial IDs versus noise. The layers of intervention are chosen as the middle third of all layers for that model. Each point shows the model's mean steerability (on x) against its accuracy on COCO-spatial with no spatial intervention. Dotted lines connect models within a family.

\textbf{Spatial Loss Module}

\label{spatial id loss}

Fig. \ref{fig:spatial ID and model performance correlation} shows spatial IDs signal stronger model performance. This suggests that the strength of spatial IDs could be a valuable learning signal for VLMs to learn principled spatial reasoning. To validate this intuition, we finetune Qwen2-2B on a synthetic dataset similar to the one used to extract spatial IDs, and evaluate on COCO-Spatial.
We introduce an additional loss module at layer 11 that computes the cosine similarity between the predicted and ground-truth spatial ID at that layer. We provide detailed explanations for this process in \S \ref{supp sec spatial loss addendum}. This spatial ID loss is added to the standard language modeling objective, providing extra supervision. We perform a control training without the spatial ID loss.  As the training, and thus the spatial alignment, was performed on simplistic synthetic data, both models saturate and start to overfit after reaching peak validation accuracy around 90\%. But we see that explicit spatial ID loss helps the model generalize faster, reaching 91\% accuracy on COCO-spatial at 3.2k steps, a net 6\% accuracy gain over the control case under the same number of training steps.

\begin{table}[h]
\centering
{
\begin{tabular}{l l c c c c c}
\toprule
& Num Steps & 0 & 800 & 1600 & 2400 & 3200 \\
\midrule
\multirow{2}{*}{\textbf{Control}}
  & LM Loss ($\downarrow$)         & 3.30 & 0.05 & \textless{}0.01 & \textless{}0.01 & \textless{}0.01 \\
  \cline{2-7}
  & COCO Val Accuracy ($\uparrow$) & 0.77 & 0.83 & 0.84            & 0.85            & \textbf{0.85} \\
\midrule
\multirow{3}{*}{\textbf{With Spatial Loss}}
  & LM Loss ($\downarrow$)         & 2.71 & 0.04 & \textless{}0.01 & \textless{}0.01 & \textless{}0.01 \\
  & Spatial ID Loss ($\downarrow$) & 0.75 & 0.58 & 0.41            & 0.36            & 0.33 \\
  \cline{2-7}
  & COCO Val Accuracy ($\uparrow$) & 0.77 & 0.83 & 0.84            & 0.88            & \textbf{0.91} \\
\bottomrule
\end{tabular}
}
\end{table}

\section{Temporal IDs in Video Models}
\label{sec video temporal IDs}

Thus far, we have characterized spatial IDs as a causal model for spatial visual reasoning in VLMs. 
Could we find a similar linear paradigm for the temporal axis? In this section, we repeat the experiments in \S \ref{sec: derive_spatial_ids}-\ref{sec:causality} for the temporal dimension in video models, with the goal of identifying linearly separable temporal markers on object words. For space, experimental procedures are described briefly here, and in greater detail in Appendix \S A.

\subsection{Mirroring, Extracting, and Steering across the Temporal Axis}

\begin{wrapfigure}{r}{0.45\textwidth}
\vspace{-4mm}
  \begin{center}
\includegraphics[width=0.45\textwidth]{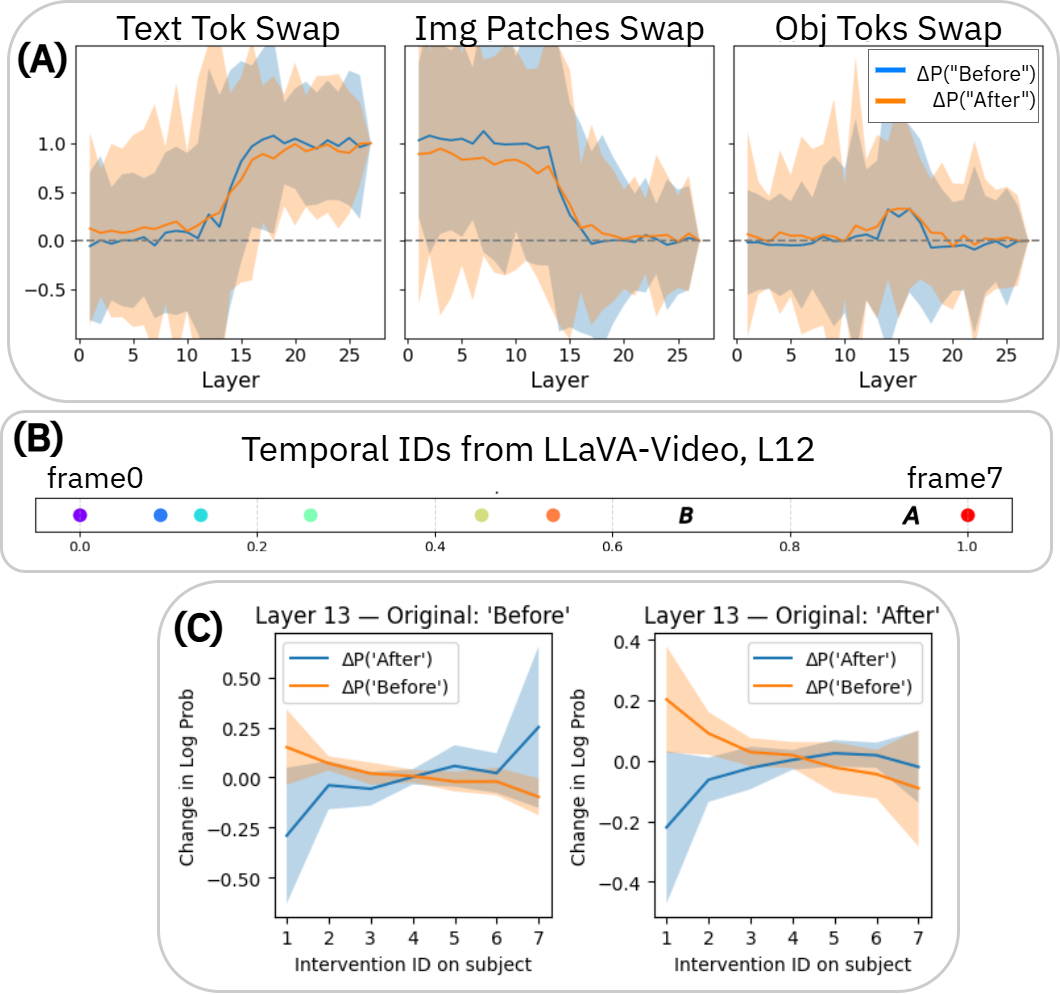}
  \end{center}
  \vspace{-5mm}
    \caption{\textbf{Temporal ID Results}. Mirror Swapping on videos (A), Temporal ID grid (B), and temporal ID steering on model beliefs (C) with \llavavid}
    \vspace{-15mm}
\label{fig:video everything temporal}
\end{wrapfigure}

\textbf{Temporal Mirror Swapping}. We validate that there exist modality alignment layers with object-level visual information transfer in video models. For mirrored videos, we take the  Scene\_QA subset 
of \textsc{MVBench} \citep{li_mvbench_2024} and swap the order of frames from back to front. Following Alg. \ref{alg:intervene_swaps_mirror}, we show results for swapping text tokens, image patches, and object words in Fig. \ref{fig:video everything temporal}A. 
While the error bound is noisier than spatial \llava, likely as LLaVA-Video follows response formats less well, we see the expected bump around middle layers for crossmodal integration.

\textbf{Temporal ID Extraction}. 
Derivation of temporal IDs and the temporal vector $t_L$ follows Eq. \ref {equation: first} - \ref{equation:direction vectors}, with synthetic 8-frame videos of \textsc{Objaverse} renders.
Results are shown in Fig. \ref{fig:video everything temporal}B. We again see that the text activation for ``before" projects closer to earlier frames, than the activation for ``after".


\textbf{Causality of Temporal IDs}. Finally, to confirm controllability with arbitrary temporal IDs, we perform the steering experiment per Alg. \ref{alg:intervene_on_obj} on \textsc{MVBench} videos. Results are shown in Fig. \ref{fig:video everything temporal}C. On these real, naturalistic videos, we see that later temporal IDs steer the model belief towards ``after", and earlier IDs towards ``before", as expected.

\subsection{Emergence of Temporal IDs}
Fig. \ref{fig:video everything temporal} shows summary results on \llavavid, but we include temporal IDs from \vidllama{} and Qwen2.5  in Appendix \S \ref{supp sec temporal grids more models}.  \llavavid {} and \vidllama{} use textual description of the video length and number of frames to indicate timestamps preceding the visual input, while Qwen uses explicit MRoPE time IDs.
This suggests that spatiotemporal IDs can emerge without explicit positional encoding, beyond the simple mechanism derived in Eq. \ref{eq:spatial-id}.

\vspace{-2mm}
\section{Related Work}
\vspace{-2mm}
\label{related_work}

Mechanistic interpretability is a growing field uncovering the inner workings of large models, popularizing techniques such as circuit tracing \citep{elhage2021mathematical, ameisen2025circuit}, Sparse Autoencoders \citep{cunningham2023sparse}, linear probing \citep{alain2016understanding}, and others. The Linear Representation Hypothesis posits that concepts are linearly encoded in LLM latents \citep{park_linear_2024}, and activation patching supports that linear changes in activations drive model belief \citep{meng2022locating, zhang2023towards}. 
Internal in-context reasoning mechanisms such as linear \textit{binding IDs}
\citep{feng_how_2024,feng_monitoring_2024} have been identified in LLMs, along with other evidence for linear multi-hop reasoning \citep{yang_large_2024}, in-context task vectors \citep{hendel_-context_2023} and linear relational embeddings \citep{hernandez_linearity_2024} during reasoning. 

Linearity of embeddings have also been discovered in VLM latent spaces \citep{trager2023linear, jiang_interpreting_2025} to some degree.
Previous work showed that VLMs separate VQA into image-focused then text-focused stages \citep{jiang_devils_2025}, and others have extended LLM interpretability techniques like logit lens \citep{neo_towards_2024} or attention tracking \citep{zhang_redundancy_2024, yu_understanding_2025} to VLMs to unearth internal circuits.
In our work, we mechanistically capture spatiotemporal information flow from image patches to text tokens in VLMs, via the spatial ID mechanism.


\vspace{-2mm}
\section{Conclusion, Limitations, \& Future Work}
\vspace{-2mm}

We propose spatiotemporal IDs as a linear model for visual reasoning about space and time in VLMs. With a series of causal analyses, we show these IDs can be obtained in many SoTA models, and that they closely mediate models' beliefs about visual objects' location in space and time. We further offer ways to extend this mechanistic insight to improving existing VLMs. 
For tractability, our work is currently limited to analyses in simple spatial queries or appearance-based temporal queries. Experimental design for more complex, open-ended queries will enhance our understanding of how VLMs utilize rudimentary concepts like spatial IDs in more nuanced settings.
Further, we only extract and steer models of sizes up to 14B parameters due to compute constraints. Investigation into whether the spatial ID circuit plays a similarly prominent role in larger models will reveal whether VLMs of varying capacities follow analogous methods for visual reasoning, or employ distinct measures. 
Lastly, while we show several potential ways to leverage spatial IDs for VLM diagnostics or finetuning, future work could include expanded use cases such as explicit temporal guidance at large scale.

\section*{Acknowledgments}
We would like to thank Aadarsh Sahoo, Jiahai Feng, Michael Hobley, and Brian Cheung for valuable discussions and proofreading.
Raphi Kang is funded by the NSF Graduate Research Fellowship. Georgia Gkioxari is supported by the Hurt Scholar program, Meta and Google. Hongqiao Chen was supported by the Caltech SURF program.

\section*{Reproducibility}
We provide finegrained details for all experiments in \S A of the Appendix, and results on all the models considered in \S B. We include experimental details, results from various ablation analyses and counterfactual trials in \S C-D. Code for all experiments can be found at:
\hyperlink{https://github.com/Raphoo/linear-mech-vlms}{https://github.com/Raphoo/linear-mech-vlms}.




\bibliography{iclr2026_conference}

@article{kamath2023s,
  title={What's" up" with vision-language models? investigating their struggle with spatial reasoning},
  author={Kamath, Amita and Hessel, Jack and Chang, Kai-Wei},
  journal={arXiv preprint arXiv:2310.19785},
  year={2023}
}

@article{ramakrishnan2018overcoming,
  title={Overcoming language priors in visual question answering with adversarial regularization},
  author={Ramakrishnan, Sainandan and Agrawal, Aishwarya and Lee, Stefan},
  journal={Advances in neural information processing systems},
  volume={31},
  year={2018}
}

@inproceedings{leng2024mitigating,
  title={Mitigating object hallucinations in large vision-language models through visual contrastive decoding},
  author={Leng, Sicong and Zhang, Hang and Chen, Guanzheng and Li, Xin and Lu, Shijian and Miao, Chunyan and Bing, Lidong},
  booktitle={Proceedings of the IEEE/CVF Conference on Computer Vision and Pattern Recognition},
  pages={13872--13882},
  year={2024}
}

@article{chen2025spatial,
  title={Why is spatial reasoning hard for vlms? an attention mechanism perspective on focus areas},
  author={Chen, Shiqi and Zhu, Tongyao and Zhou, Ruochen and Zhang, Jinghan and Gao, Siyang and Niebles, Juan Carlos and Geva, Mor and He, Junxian and Wu, Jiajun and Li, Manling},
  journal={arXiv preprint arXiv:2503.01773},
  year={2025}
}

@inproceedings{tong2024eyes,
  title={Eyes wide shut? exploring the visual shortcomings of multimodal llms},
  author={Tong, Shengbang and Liu, Zhuang and Zhai, Yuexiang and Ma, Yi and LeCun, Yann and Xie, Saining},
  booktitle={Proceedings of the IEEE/CVF Conference on Computer Vision and Pattern Recognition},
  pages={9568--9578},
  year={2024}
}

@article{stogiannidis2025mind,
  title={Mind the gap: Benchmarking spatial reasoning in vision-language models},
  author={Stogiannidis, Ilias and McDonagh, Steven and Tsaftaris, Sotirios A},
  journal={arXiv preprint arXiv:2503.19707},
  year={2025}
}

@inproceedings{chen2024spatialvlm,
  title={Spatialvlm: Endowing vision-language models with spatial reasoning capabilities},
  author={Chen, Boyuan and Xu, Zhuo and Kirmani, Sean and Ichter, Brain and Sadigh, Dorsa and Guibas, Leonidas and Xia, Fei},
  booktitle={Proceedings of the IEEE/CVF Conference on Computer Vision and Pattern Recognition},
  pages={14455--14465},
  year={2024}
}

@inproceedings{xiao2024can,
  title={Can i trust your answer? visually grounded video question answering},
  author={Xiao, Junbin and Yao, Angela and Li, Yicong and Chua, Tat-Seng},
  booktitle={Proceedings of the IEEE/CVF Conference on Computer Vision and Pattern Recognition},
  pages={13204--13214},
  year={2024}
}

@inproceedings{lin2014microsoft,
  title={Microsoft coco: Common objects in context},
  author={Lin, Tsung-Yi and Maire, Michael and Belongie, Serge and Hays, James and Perona, Pietro and Ramanan, Deva and Doll{\'a}r, Piotr and Zitnick, C Lawrence},
  booktitle={European conference on computer vision},
  pages={740--755},
  year={2014},
  organization={Springer}
}

@misc{feng_how_2024,
	title = {How do {Language} {Models} {Bind} {Entities} in {Context}?},
	url = {http://arxiv.org/abs/2310.17191},
	doi = {10.48550/arXiv.2310.17191},
	urldate = {2025-01-27},
	publisher = {arXiv},
	author = {Feng, Jiahai and Steinhardt, Jacob},
	month = may,
	year = {2024},
	note = {arXiv:2310.17191 [cs]},
}

@misc{liu_visual_2023,
	title = {Visual {Instruction} {Tuning}},
	url = {http://arxiv.org/abs/2304.08485},
	doi = {10.48550/arXiv.2304.08485},
	urldate = {2025-02-11},
	publisher = {arXiv},
	author = {Liu, Haotian and Li, Chunyuan and Wu, Qingyang and Lee, Yong Jae},
	month = dec,
	year = {2023},
	note = {arXiv:2304.08485 [cs]},
}

@misc{yu_understanding_2025,
	title = {Understanding {Multimodal} {LLMs}: the {Mechanistic} {Interpretability} of {Llava} in {Visual} {Question} {Answering}},
	shorttitle = {Understanding {Multimodal} {LLMs}},
	url = {http://arxiv.org/abs/2411.10950},
	doi = {10.48550/arXiv.2411.10950},
	urldate = {2025-03-12},
	author = {Yu, Zeping and Ananiadou, Sophia},
	month = jan,
	year = {2025},
	note = {arXiv:2411.10950 [cs]},
}

@misc{neo_towards_2024,
	title = {Towards {Interpreting} {Visual} {Information} {Processing} in {Vision}-{Language} {Models}},
	url = {http://arxiv.org/abs/2410.07149},
	doi = {10.48550/arXiv.2410.07149},
	urldate = {2025-03-13},
	publisher = {arXiv},
	author = {Neo, Clement and Ong, Luke and Torr, Philip and Geva, Mor and Krueger, David and Barez, Fazl},
	month = oct,
	year = {2024},
	note = {arXiv:2410.07149 [cs]},
}

@misc{feng_monitoring_2024,
	title = {Monitoring {Latent} {World} {States} in {Language} {Models} with {Propositional} {Probes}},
	url = {http://arxiv.org/abs/2406.19501},
	doi = {10.48550/arXiv.2406.19501},
	urldate = {2025-03-17},
	publisher = {arXiv},
	author = {Feng, Jiahai and Russell, Stuart and Steinhardt, Jacob},
	month = dec,
	year = {2024},
	note = {arXiv:2406.19501 [cs]},
}

@misc{jiang_interpreting_2025,
	title = {Interpreting and {Editing} {Vision}-{Language} {Representations} to {Mitigate} {Hallucinations}},
	url = {http://arxiv.org/abs/2410.02762},
	doi = {10.48550/arXiv.2410.02762},
	urldate = {2025-03-17},
	publisher = {arXiv},
	author = {Jiang, Nick and Kachinthaya, Anish and Petryk, Suzie and Gandelsman, Yossi},
	month = feb,
	year = {2025},
	note = {arXiv:2410.02762 [cs]},
}

@misc{jiang_devils_2025,
	title = {Devils in {Middle} {Layers} of {Large} {Vision}-{Language} {Models}: {Interpreting}, {Detecting} and {Mitigating} {Object} {Hallucinations} via {Attention} {Lens}},
	shorttitle = {Devils in {Middle} {Layers} of {Large} {Vision}-{Language} {Models}},
	url = {http://arxiv.org/abs/2411.16724},
	doi = {10.48550/arXiv.2411.16724},
	urldate = {2025-04-11},
	publisher = {arXiv},
	author = {Jiang, Zhangqi and Chen, Junkai and Zhu, Beier and Luo, Tingjin and Shen, Yankun and Yang, Xu},
	month = apr,
	year = {2025},
	note = {arXiv:2411.16724 [cs]},
}

@misc{hernandez_linearity_2024,
	title = {Linearity of {Relation} {Decoding} in {Transformer} {Language} {Models}},
	url = {http://arxiv.org/abs/2308.09124},
	doi = {10.48550/arXiv.2308.09124},
	urldate = {2025-04-26},
	author = {Hernandez, Evan and Sharma, Arnab Sen and Haklay, Tal and Meng, Kevin and Wattenberg, Martin and Andreas, Jacob and Belinkov, Yonatan and Bau, David},
	month = feb,
	year = {2024},
	note = {arXiv:2308.09124 [cs]},
}

@misc{zhang_videollama_2025,
	title = {{VideoLLaMA} 3: {Frontier} {Multimodal} {Foundation} {Models} for {Image} and {Video} {Understanding}},
	shorttitle = {{VideoLLaMA} 3},
	url = {http://arxiv.org/abs/2501.13106},
	doi = {10.48550/arXiv.2501.13106},
	urldate = {2025-05-06},
	publisher = {arXiv},
	author = {Zhang, Boqiang and Li, Kehan and Cheng, Zesen and Hu, Zhiqiang and Yuan, Yuqian and Chen, Guanzheng and Leng, Sicong and Jiang, Yuming and Zhang, Hang and Li, Xin and Jin, Peng and Zhang, Wenqi and Wang, Fan and Bing, Lidong and Zhao, Deli},
	month = jan,
	year = {2025},
	note = {arXiv:2501.13106 [cs]},
}

@misc{zhang_video_2024,
	title = {Video {Instruction} {Tuning} {With} {Synthetic} {Data}},
	url = {http://arxiv.org/abs/2410.02713},
	doi = {10.48550/arXiv.2410.02713},
	urldate = {2025-05-06},
	publisher = {arXiv},
	author = {Zhang, Yuanhan and Wu, Jinming and Li, Wei and Li, Bo and Ma, Zejun and Liu, Ziwei and Li, Chunyuan},
	month = oct,
	year = {2024},
	note = {arXiv:2410.02713 [cs]},
}

@inproceedings{vaswani2017attention,
  title     = {Attention Is All You Need},
  author    = {Vaswani, Ashish and Shazeer, Noam and Parmar, Niki and Uszkoreit, Jakob and Jones, Llion and Gomez, Aidan N. and Kaiser, {\L}ukasz and Polosukhin, Illia},
  booktitle = {Advances in Neural Information Processing Systems (NeurIPS)},
  year      = {2017},
  archivePrefix = {arXiv},
  eprint    = {1706.03762},
  url       = {https://arxiv.org/abs/1706.03762}
}

@inproceedings{li_mvbench_2024,
	address = {Seattle, WA, USA},
	title = {{MVBench}: {A} {Comprehensive} {Multi}-modal {Video} {Understanding} {Benchmark}},
	copyright = {https://doi.org/10.15223/policy-029},
	isbn = {9798350353006},
	shorttitle = {{MVBench}},
	url = {https://ieeexplore.ieee.org/document/10658165/},
	doi = {10.1109/CVPR52733.2024.02095},
	language = {en},
	urldate = {2025-05-08},
	booktitle = {2024 {IEEE}/{CVF} {Conference} on {Computer} {Vision} and {Pattern} {Recognition} ({CVPR})},
	publisher = {IEEE},
	author = {Li, Kunchang and Wang, Yali and He, Yinan and Li, Yizhuo and Wang, Yi and Liu, Yi and Wang, Zun and Xu, Jilan and Chen, Guo and Lou, Ping and Wang, Limin and Qiao, Yu},
	month = jun,
	year = {2024},
	pages = {22195--22206},
}

@misc{zhang_redundancy_2024,
	title = {From {Redundancy} to {Relevance}: {Information} {Flow} in {LVLMs} {Across} {Reasoning} {Tasks}},
	shorttitle = {From {Redundancy} to {Relevance}},
	url = {http://arxiv.org/abs/2406.06579},
	doi = {10.48550/arXiv.2406.06579},
	urldate = {2025-05-27},
	publisher = {arXiv},
	author = {Zhang, Xiaofeng and Quan, Yihao and Shen, Chen and Yuan, Xiaosong and Yan, Shaotian and Xie, Liang and Wang, Wenxiao and Gu, Chaochen and Tang, Hao and Ye, Jieping},
	month = oct,
	year = {2024},
	note = {arXiv:2406.06579 [cs]},
}

@misc{yang_large_2024,
	title = {Do {Large} {Language} {Models} {Latently} {Perform} {Multi}-{Hop} {Reasoning}?},
	url = {http://arxiv.org/abs/2402.16837},
	doi = {10.48550/arXiv.2402.16837},
	urldate = {2025-05-28},
	publisher = {arXiv},
	author = {Yang, Sohee and Gribovskaya, Elena and Kassner, Nora and Geva, Mor and Riedel, Sebastian},
	month = feb,
	year = {2024},
	note = {arXiv:2402.16837 [cs]},
}

@misc{hendel_-context_2023,
	title = {In-{Context} {Learning} {Creates} {Task} {Vectors}},
	url = {http://arxiv.org/abs/2310.15916},
	doi = {10.48550/arXiv.2310.15916},
	urldate = {2025-05-28},
	publisher = {arXiv},
	author = {Hendel, Roee and Geva, Mor and Globerson, Amir},
	month = oct,
	year = {2023},
	note = {arXiv:2310.15916 [cs]},
}

@misc{park_linear_2024,
	title = {The {Linear} {Representation} {Hypothesis} and the {Geometry} of {Large} {Language} {Models}},
	url = {http://arxiv.org/abs/2311.03658},
	doi = {10.48550/arXiv.2311.03658},
	urldate = {2025-06-03},
	publisher = {arXiv},
	author = {Park, Kiho and Choe, Yo Joong and Veitch, Victor},
	month = jul,
	year = {2024},
	note = {arXiv:2311.03658 [cs]},
}

@misc{bai_qwen25-vl_2025,
	title = {Qwen2.5-{VL} {Technical} {Report}},
	url = {http://arxiv.org/abs/2502.13923},
	doi = {10.48550/arXiv.2502.13923},
	urldate = {2025-06-09},
	publisher = {arXiv},
	author = {Bai, Shuai and Chen, Keqin and Liu, Xuejing and Wang, Jialin and Ge, Wenbin and Song, Sibo and Dang, Kai and Wang, Peng and Wang, Shijie and Tang, Jun and Zhong, Humen and Zhu, Yuanzhi and Yang, Mingkun and Li, Zhaohai and Wan, Jianqiang and Wang, Pengfei and Ding, Wei and Fu, Zheren and Xu, Yiheng and Ye, Jiabo and Zhang, Xi and Xie, Tianbao and Cheng, Zesen and Zhang, Hang and Yang, Zhibo and Xu, Haiyang and Lin, Junyang},
	month = feb,
	year = {2025},
	note = {arXiv:2502.13923 [cs]},
}

@misc{fan_grit_2025,
	title = {{GRIT}: {Teaching} {MLLMs} to {Think} with {Images}},
	shorttitle = {{GRIT}},
	url = {http://arxiv.org/abs/2505.15879},
	doi = {10.48550/arXiv.2505.15879},
	urldate = {2025-06-09},
	publisher = {arXiv},
	author = {Fan, Yue and He, Xuehai and Yang, Diji and Zheng, Kaizhi and Kuo, Ching-Chen and Zheng, Yuting and Narayanaraju, Sravana Jyothi and Guan, Xinze and Wang, Xin Eric},
	month = may,
	year = {2025},
	note = {arXiv:2505.15879 [cs]},
}

@misc{chen_internvl_2024,
	title = {{InternVL}: {Scaling} up {Vision} {Foundation} {Models} and {Aligning} for {Generic} {Visual}-{Linguistic} {Tasks}},
	shorttitle = {{InternVL}},
	url = {http://arxiv.org/abs/2312.14238},
	doi = {10.48550/arXiv.2312.14238},
	urldate = {2025-06-11},
	publisher = {arXiv},
	author = {Chen, Zhe and Wu, Jiannan and Wang, Wenhai and Su, Weijie and Chen, Guo and Xing, Sen and Zhong, Muyan and Zhang, Qinglong and Zhu, Xizhou and Lu, Lewei and Li, Bin and Luo, Ping and Lu, Tong and Qiao, Yu and Dai, Jifeng},
	month = jan,
	year = {2024},
	note = {arXiv:2312.14238 [cs]},
}

@article{ameisen2025circuit,
  author={Ameisen, Emmanuel and Lindsey, Jack and Pearce, Adam and Gurnee, Wes and Turner, Nicholas L. and Chen, Brian and Citro, Craig and Abrahams, David and Carter, Shan and Hosmer, Basil and Marcus, Jonathan and Sklar, Michael and Templeton, Adly and Bricken, Trenton and McDougall, Callum and Cunningham, Hoagy and Henighan, Thomas and Jermyn, Adam and Jones, Andy and Persic, Andrew and Qi, Zhenyi and Ben Thompson, T. and Zimmerman, Sam and Rivoire, Kelley and Conerly, Thomas and Olah, Chris and Batson, Joshua},
  title={Circuit Tracing: Revealing Computational Graphs in Language Models},
  journal={Transformer Circuits Thread},
  year={2025},
  url={https://transformer-circuits.pub/2025/attribution-graphs/methods.html}
}

@inproceedings{ren2023masked,
  title={Masked jigsaw puzzle: A versatile position embedding for vision transformers},
  author={Ren, Bin and Liu, Yahui and Song, Yue and Bi, Wei and Cucchiara, Rita and Sebe, Nicu and Wang, Wei},
  booktitle={Proceedings of the IEEE/CVF Conference on Computer Vision and Pattern Recognition},
  pages={20382--20391},
  year={2023}
}

@article{alain2016understanding,
  title={Understanding intermediate layers using linear classifier probes},
  author={Alain, Guillaume and Bengio, Yoshua},
  journal={arXiv preprint arXiv:1610.01644},
  year={2016}
}

@article{team2024gemma,
  title={Gemma: Open models based on gemini research and technology},
  author={Team, Gemma and Mesnard, Thomas and Hardin, Cassidy and Dadashi, Robert and Bhupatiraju, Surya and Pathak, Shreya and Sifre, Laurent and Rivi{\`e}re, Morgane and Kale, Mihir Sanjay and Love, Juliette and others},
  journal={arXiv preprint arXiv:2403.08295},
  year={2024}
}

@article{cunningham2023sparse,
  title={Sparse autoencoders find highly interpretable features in language models},
  author={Cunningham, Hoagy and Ewart, Aidan and Riggs, Logan and Huben, Robert and Sharkey, Lee},
  journal={arXiv preprint arXiv:2309.08600},
  year={2023}
}

@article{elhage2021mathematical,
   title={A Mathematical Framework for Transformer Circuits},
   author={Elhage, Nelson and Nanda, Neel and Olsson, Catherine and Henighan, Tom and Joseph, Nicholas and Mann, Ben and Askell, Amanda and Bai, Yuntao and Chen, Anna and Conerly, Tom and DasSarma, Nova and Drain, Dawn and Ganguli, Deep and Hatfield-Dodds, Zac and Hernandez, Danny and Jones, Andy and Kernion, Jackson and Lovitt, Liane and Ndousse, Kamal and Amodei, Dario and Brown, Tom and Clark, Jack and Kaplan, Jared and McCandlish, Sam and Olah, Chris},
   year={2021},
   journal={Transformer Circuits Thread},
   note={https://transformer-circuits.pub/2021/framework/index.html}
}

@inproceedings{trager2023linear,
  title={Linear spaces of meanings: compositional structures in vision-language models},
  author={Trager, Matthew and Perera, Pramuditha and Zancato, Luca and Achille, Alessandro and Bhatia, Parminder and Soatto, Stefano},
  booktitle={Proceedings of the IEEE/CVF International Conference on Computer Vision},
  pages={15395--15404},
  year={2023}
}

@article{zhang2023towards,
  title={Towards best practices of activation patching in language models: Metrics and methods},
  author={Zhang, Fred and Nanda, Neel},
  journal={arXiv preprint arXiv:2309.16042},
  year={2023}
}

@article{meng2022locating,
  title={Locating and editing factual associations in gpt},
  author={Meng, Kevin and Bau, David and Andonian, Alex and Belinkov, Yonatan},
  journal={Advances in neural information processing systems},
  volume={35},
  pages={17359--17372},
  year={2022}
}

@article{li2024temporal,
  title={Temporal reasoning transfer from text to video},
  author={Li, Lei and Liu, Yuanxin and Yao, Linli and Zhang, Peiyuan and An, Chenxin and Wang, Lean and Sun, Xu and Kong, Lingpeng and Liu, Qi},
  journal={arXiv preprint arXiv:2410.06166},
  year={2024}
}

@article{mcknight2010mann,
  title={Mann-whitney U test},
  author={McKnight, Patrick E and Najab, Julius},
  journal={The Corsini encyclopedia of psychology},
  pages={1--1},
  year={2010},
  publisher={Wiley Online Library}
}

@article{kang2025clip,
  title={Is CLIP ideal? No. Can we fix it? Yes!},
  author={Kang, Raphi and Song, Yue and Gkioxari, Georgia and Perona, Pietro},
  journal={arXiv preprint arXiv:2503.08723},
  year={2025}
}

@inproceedings{petsiuk2021black,
  title={Black-box explanation of object detectors via saliency maps},
  author={Petsiuk, Vitali and Jain, Rajiv and Manjunatha, Varun and Morariu, Vlad I and Mehra, Ashutosh and Ordonez, Vicente and Saenko, Kate},
  booktitle={Proceedings of the IEEE/CVF conference on computer vision and pattern recognition},
  pages={11443--11452},
  year={2021}
}

@inproceedings{deitke2023objaverse,
  title={Objaverse: A universe of annotated 3d objects},
  author={Deitke, Matt and Schwenk, Dustin and Salvador, Jordi and Weihs, Luca and Michel, Oscar and VanderBilt, Eli and Schmidt, Ludwig and Ehsani, Kiana and Kembhavi, Aniruddha and Farhadi, Ali},
  booktitle={Proceedings of the IEEE/CVF conference on computer vision and pattern recognition},
  pages={13142--13153},
  year={2023}
}

@article{dubey2024llama,
  title={The llama 3 herd of models},
  author={Dubey, Abhimanyu and Jauhri, Abhinav and Pandey, Abhinav and Kadian, Abhishek and Al-Dahle, Ahmad and Letman, Aiesha and Mathur, Akhil and Schelten, Alan and Yang, Amy and Fan, Angela and others},
  journal={arXiv e-prints},
  pages={arXiv--2407},
  year={2024}
}
\bibliographystyle{iclr2026_conference}

\newpage

\appendix
\startcontents[appendices]
\printcontents[appendices]{l}{1}{\section*{Linear Mechanisms for Spatiotemporal Reasoning in Vision Language Models\\--Supplementary Material--}}


    \renewcommand{\thefigure}{A\arabic{figure}}
    \setcounter{figure}{0}
\vspace{15mm}

\section{Experimental Details}

\subsection{Mirror Swapping}
\label{appendix:mirror swapping section}

\begin{figure}[H]
    \centering
    \includegraphics[width=0.55\linewidth]{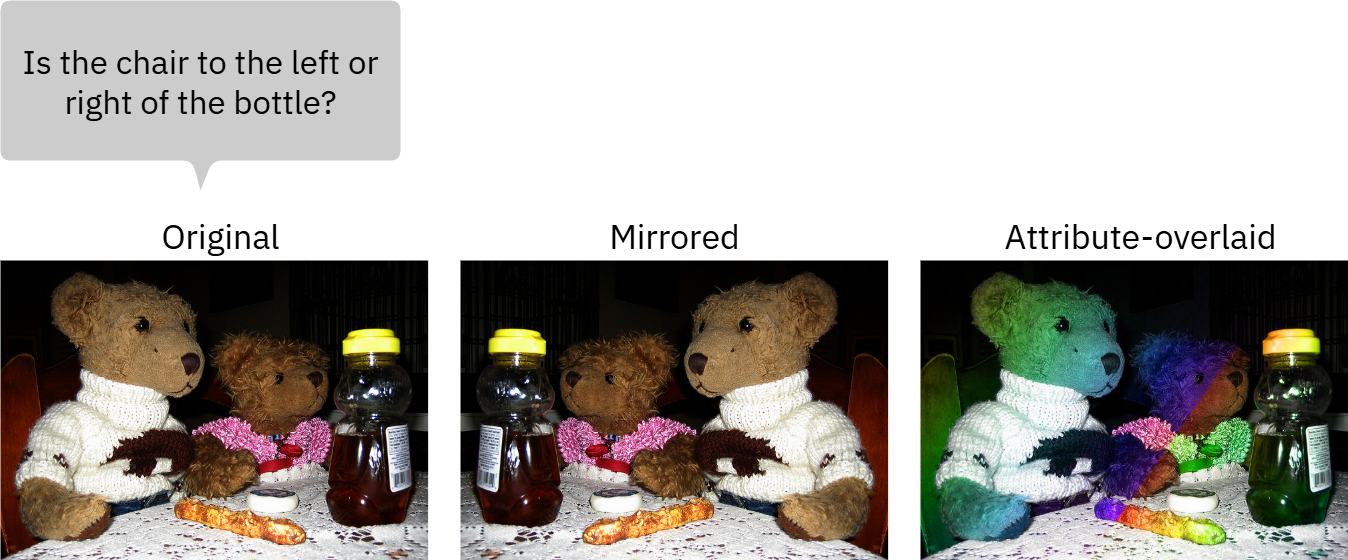}
    \caption{Example altered images for mirror swapping and attribute swapping.}
    \label{fig:mirror swap supp}
\end{figure}

\textbf{Token handling}. 
Different models and tokenizers have different tokenizing schemes. For example, for the query ``Question: Is the the thermometer to the left or right of the desktop? Answer left or right. Answer: ", the tokenization from Gemma, LLaMA, LLaVA, and Qwen will be as shown:

\begin{spverbatim}
=== Gemma ===
Tokens: ['Question', ':', '▁Is', '▁the', '▁thermometer', '▁to', '▁the', '▁left', '▁or', '▁right', '▁of', '▁the', '▁desktop', '?', '▁Answer', '▁left', '▁or', '▁right', '.', '▁Answer', ':']

=== LLaMA ===
Tokens: ['▁Question', ':', '▁Is', '▁the', '▁therm', 'ometer', '▁to', '▁the', '▁left', '▁or', '▁right', '▁of', '▁the', '▁desktop', '?', '▁Answer', '▁left', '▁or', '▁right', '.', '▁Answer', ':']

=== LLaVA ===
Tokens: ['▁Question', ':', '▁Is', '▁the', '▁therm', 'ometer', '▁to', '▁the', '▁left', '▁or', '▁right', '▁of', '▁the', '▁desktop', '?', '▁Answer', '▁left', '▁or', '▁right', '.', '▁Answer', ':']

=== Qwen ===
Tokens: ['Question', ':', 'ĠIs', 'Ġthe', 'Ġthermometer', 'Ġto', 'Ġthe', 'Ġleft', 'Ġor', 'Ġright', 'Ġof', 'Ġthe', 'Ġdesktop', '?', 'ĠAnswer', 'Ġleft', 'Ġor', 'Ġright', '.', 'ĠAnswer', ':']
\end{spverbatim}

When a word is represented as multiple tokens per a model's processor (e.g., $frog$ is tokenized into [$\_f, rog$] in LLaVA), we take the last index of this list to be most representative of the object, as it is the distinguishing element. So in the case of LLaMA or LLaVA, we would take the ``ometer" token to represent the object ``thermometer".

\textbf{Logit Probabilities}. For assessing the model's likelihood of saying ``left" vs. ``right", or two other options, we take the log probability for that token following the tokenization scheme of the model family. This means we take the model output.logits and index at the token ID for 'Ġright' in Qwen, for example, to get P(``right"). 

\textbf{Activation Patching}. For every model, we first register a forward hook at each layer to collect the intermediate activation for both the original (case 1) and mirror-swapped (case 2) cases. Then, we register another forward-hook replace the original activations with the mirror-swapped one at select indices according to the three different settings.

\newpage 
\subsection{Spatiotemporal ID Extraction}
\label{appendix sec: spatial ID extract instruct}

\begin{figure}[H]
    \centering
    \includegraphics[width=0.67\linewidth]{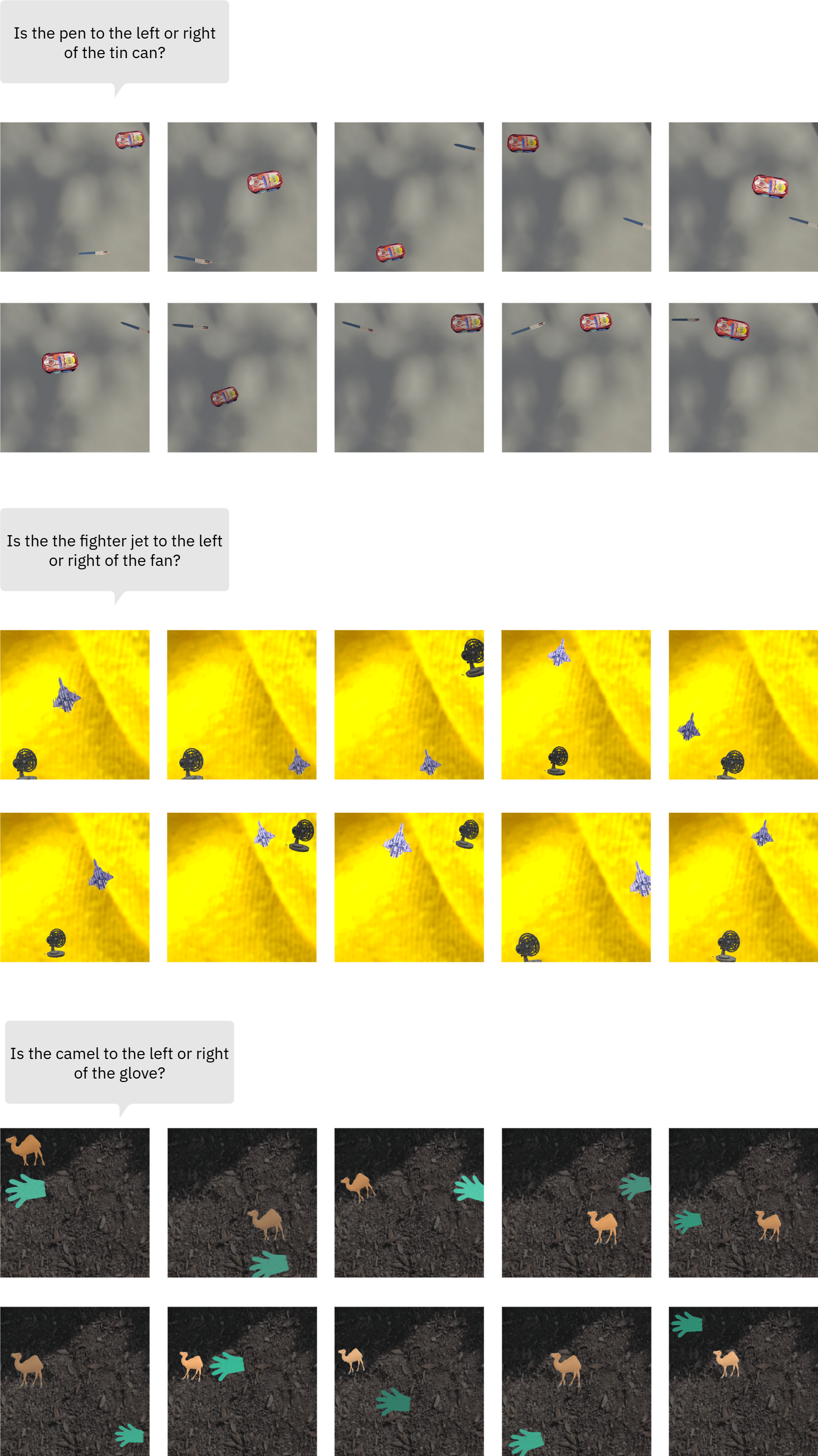}

    \caption{Synthetic images used towards spatial ID extraction}
    \label{fig:imgs for synth app}
\end{figure}

\begin{figure}[H]
    \centering
    \includegraphics[width=0.8\linewidth]{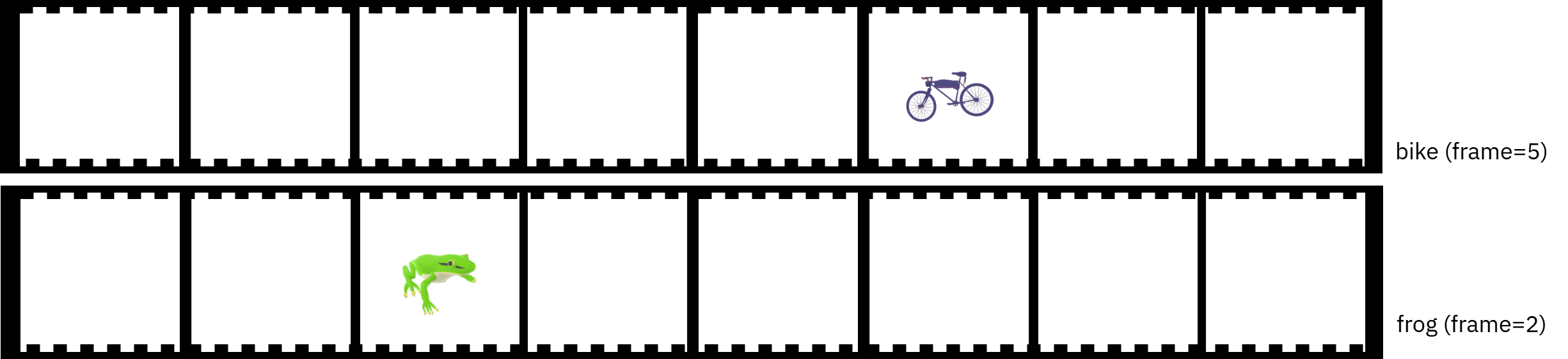}
    \caption{Illustration of synthetic videos used towards temporal ID generation. All videos had 8 frames.}
    \label{fig:vids for synth app}
\end{figure}

\textbf{Synthetic Image generation for Spatial IDs}.
We use 55 \textsc{Objaverse} object renders  and project them in various pairs onto random backgrounds, per \citep{kang2025clip}. All images with the same objects get the same text query. In \S \ref{scaling supp sec} we show the different number of objects and total number of images used to generate spatial IDs.  For $w$ object pairs, we generate $w \times s \times m^2 \times(m^2 - 1)$ total images, where $s$ is the number of object sizes we consider, and $m$ is grid size.
While we find minimal difference with extraction dataset size, as shown in \S \ref{scaling supp sec}, we use 90 object pairs, and consider $s = 4$ from \{224, 174, 124, 74\}, yielding 86,400 images. Note that each image size is $224 \times 4 = 896$ in width and height.

\textbf{Synthetic Video generation for Temporal IDs}. 
We take 5 unique \textsc{Objaverse} object pairs in 61 distinct temporal arrangements.
For baseline temporal ID extraction, all objects were centered in the image. For spatial vs. temporal disentanglement verification, we try three spatial variants - left, center, and right - for object location.

\begin{figure}[h]
    \centering
\includegraphics[width=1.0\linewidth]{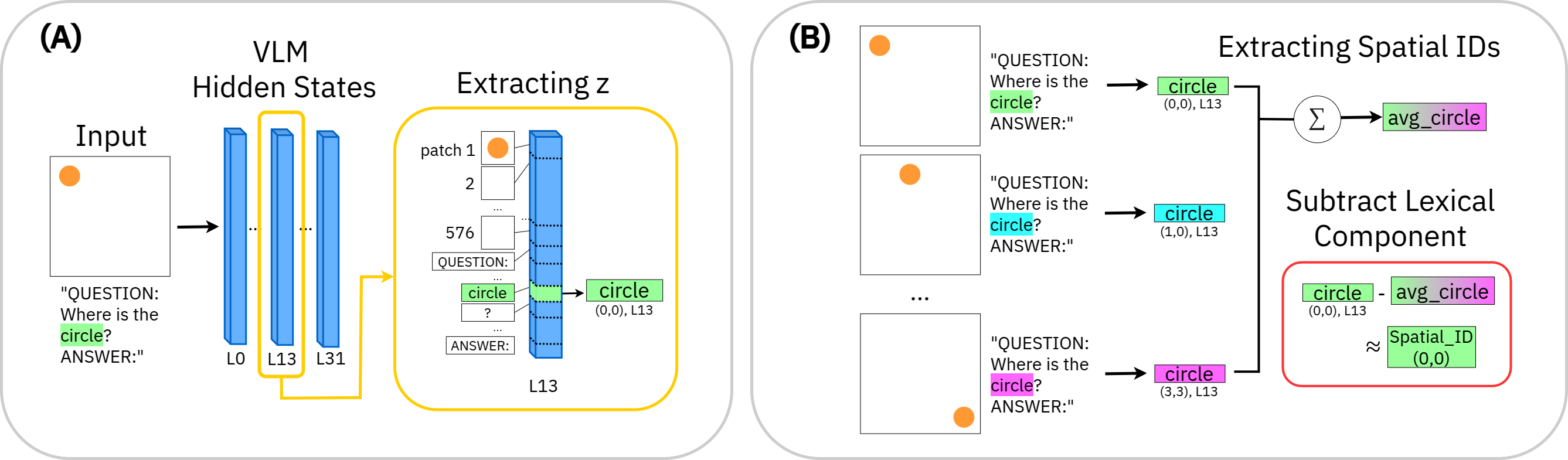}\vspace{-3mm}
\caption{Illustration of spatial ID extraction. We isolate the relevant visual object word token in a chosen layer activation (A) and compute the shared lexical component for that particular object word that is independent of spatial localization (B) to acquire the linearly bound spatial ID.}
    \label{fig:spatial-id-extract}
\end{figure}

\subsection{Arbitrary Steering Experiments}
\label{appendix sec: steering exp instruct}

\begin{figure}[H]
    \centering
    \includegraphics[width=0.85\linewidth]{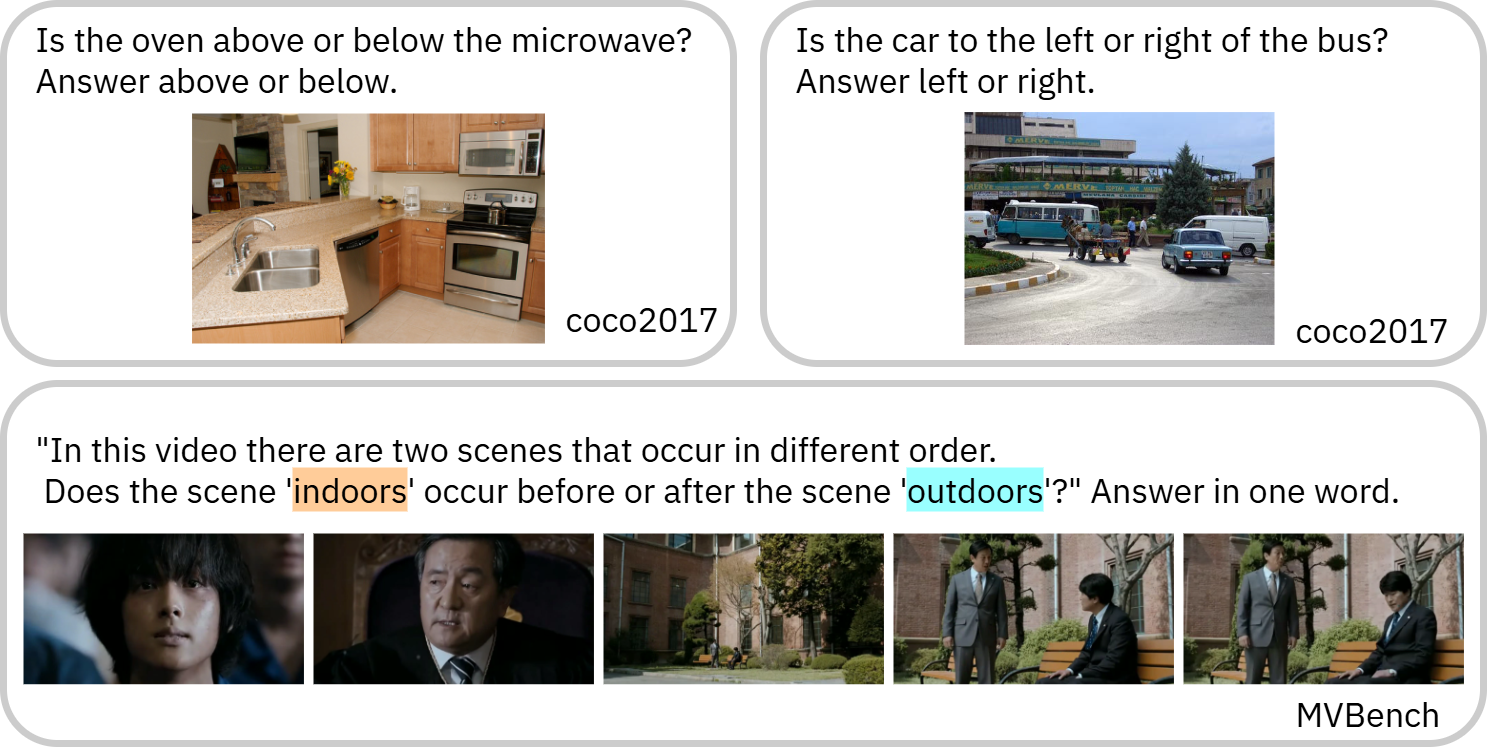}
    \caption{Examples of real images (top) and videos (bottom) we use to test model beliefs.}
    \label{fig:steering examples appendix}
\end{figure}





\subsection{Color-binding Reasoning Experiments}

\begin{figure}[h]
    \centering
    \includegraphics[width=0.89\linewidth]{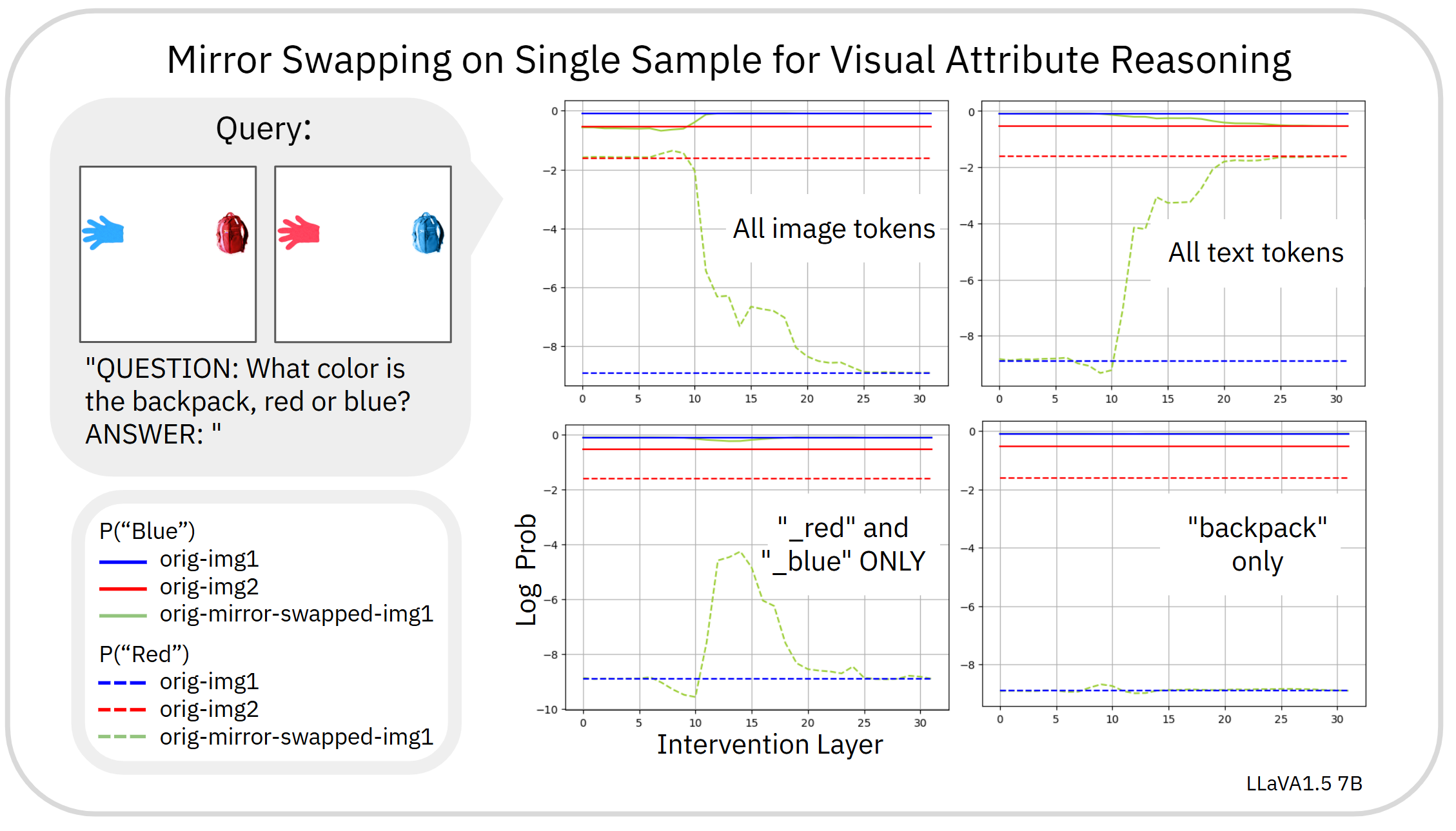}
    \caption{Mirror Swapping on Single Sample for Color Binding}
    \label{fig:single sample color bind}
\end{figure}

Do spatial IDs mediate visual reasoning beyond direct spatial queries (such as A above/below B, etc.)? To test this, we perform mirror swapping on two images where two objects are \textit{in the same place}, unlike the mirror swapping in \S \ref{sec information flow}. This time, the objects are opposing in color. 
Fig. \ref{fig:single sample color bind} shows the example query setup, as well as the results of swapping all image tokens, all text tokens, just the color word tokens, or just the object word token.

Notice that swapping the activations for “backpack” has no effect, since the spatial ID encoded in the object word activation stays the same regardless of the input image (the backpack is in the same location in both images). Swapping the activations of color-related words, on the other hand, alters model belief at key modality binding layers. This suggests that the color words were storing spatial IDs that corresponded to the location where that color was present, and matching this color spatial ID to the object token was the readout process.

\begin{figure}[h]
    \centering
    \includegraphics[width=0.89\linewidth]{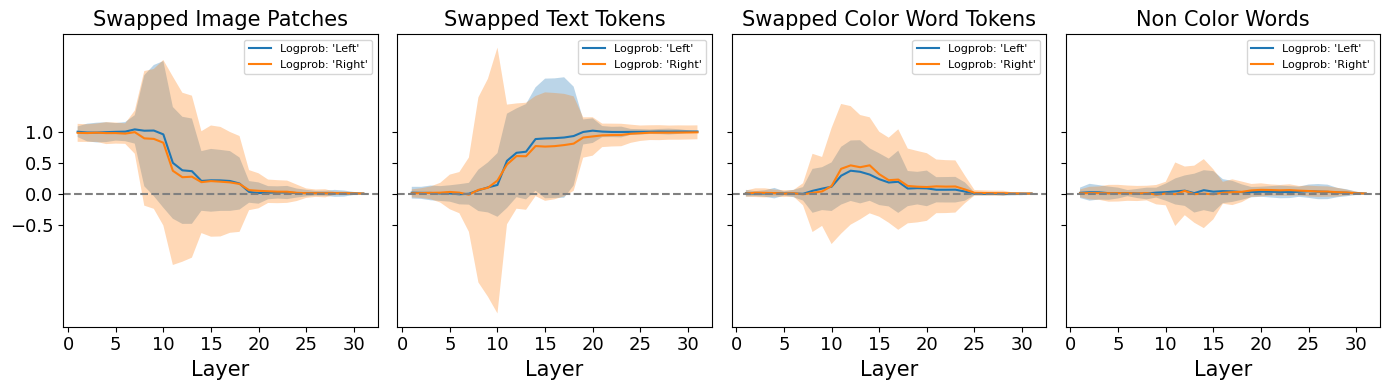}
    \caption{Swapping tokens for color binding.}
    \label{fig:avg color swap}
\end{figure}

We repeat this experiment across 100 total such images, and show the results in Fig. \ref{fig:avg color swap}. On average we see that swapping the color word tokens influences model beliefs in intermediate layers, much more so than swapping non-color word tokens. This suggests that spatial IDs mediate visual reasoning beyond direct spatial queries.

\subsection{Adversarial Steering Experiments}
\label{appendix sec: adversarial steering fig 2}

We perform steering on layers 9 through layer 2(model\_len // 3) per model. 
To ensure activation norms don't explode, we test a few different scaling factors for the interevning spatial ID's norm. In the scaling factor = 1 case, we scale the norm of the spatial ID to equal the norm of that word token's activation vector. We try a few scaling factors and choose 5 for steering all models, both for the noise vectors as well as the spatial IDs.
Here, the norm of the spatial ID is fine to exceed that of the original token activation, as we subtract the opposing spatial ID to readjust the norm. This is shown in Alg. \ref{alg:intervene_on_obj}. For confidence intervals, we choose the three layers which had greatest steering effect, and report equivalent layers' effects for the noise case.

\subsection{ID Deviation}

\textbf{Classifying Model Belief.} For the ID deviation experiment, we classified the model's belief based on its decoded response. If the response contains only the correct spatial relationship (e.g. left) and not the incorrect spatial relationship, it's considered correct. If the response contains only the incorrect spatial relationship, it's considered incorrect. If none or both were present, it's considered nonsense and discarded.


\subsection{Obfuscating Experiments}

We take images from \textsc{COCO\_spatial} and Gaussian blur different regions, as below. We generate 4 images blurred in incorrect regions in addition to the 1 image with the correct bounding box blurred. For the sensitivity, we take the difference between the outside region which changed the model belief the most, and the bounding box.

\begin{figure}[h]
    \centering
    \includegraphics[width=0.7\linewidth]{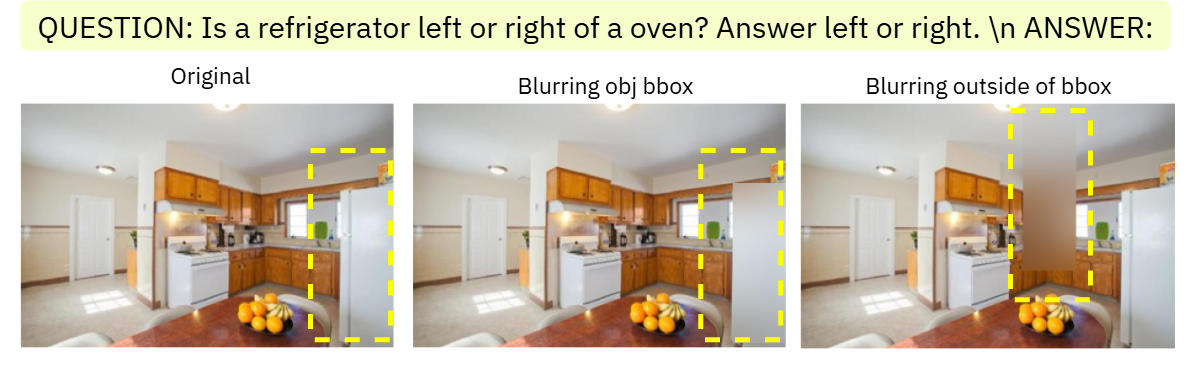}
    \caption{Example of original query and two blurred options. Yellow grid lines are just for visualization.}
    \label{fig:blurring examples}
\end{figure}

\subsection{Model Diagnosis Addendum}

\label{supp sec: model diagnosis addendum }

\textbf{Oracle Injection Experiment}
To further isolate what model components may be responsible for creating incorrect spatial IDs, we conduct the oracle injection experiment. Specifically, we intervene with the \textit{correct} spatial IDs on the object words at different layers, and see how that changes model accuracy from the control case without any intervention.

\begin{figure}[h]
    \centering
    \includegraphics[width=0.55\linewidth]{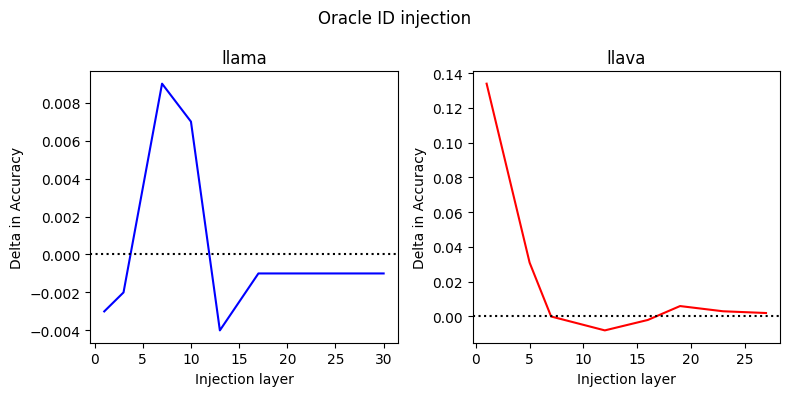}
    \caption{LLaMA and LLaVA evaluation accuracy on synthetic grid-like data with oracle spatial ID injections at varying layers. 0 is baseline model performance, without any intervention.}
    \label{fig:oracle}
\end{figure}

In accordance with our preliminary conclusion from \S \ref{sec diagnosis}, we see that LLaVA models' accuracy increases 13.4\% above the baseline when injected with oracle truth spatial IDs at layer 1. This suggests that indeed, if the image encoder had supplied correct spatial information, the downstream LM of LLaVA would have yielded greater accuracy. Intervention on intermediate to later layers in LLaVA has little effect. In LLaMA, we see that intervening on the earliest layers actually has little effect, while intervening on intermediate layers preceding the modality integration layer increases model accuracy by a modest amount (~1\%). Note that the low percentage is likely because LLaMA has higher accuracy on this spatial dataset to begin with. This behavior is in line with our expectation from \S \ref{sec diagnosis}, where we do not expect it to benefit greatly from altering image encoder spatial localization performance, but instead benefit from spatial information condensation into the proper tokens.

For this experiment, we evaluated on synthetic images made with objaverse, such as those shown in Fig. \ref{fig:imgs for synth app}. The interventions were performed with IDs from layer 17 on LLaMA for all layers including and below 17, and IDs from layer 12 on LLaVA for all layers including and below 12, as these were the layers identified as carrying spatial ID information in these respective models.
LLaMA interventions were performed at layers [1, 3, 7, 10, 13, 17, 21, 25, 30, 35] and LLaVA interventions on [1, 5, 7, 12, 16, 19, 23, 27].

 \subsection{Model finetuning with Spatial Loss}
\label{supp sec spatial loss addendum}

\textbf{Spatial ID Loss Module} In \S \ref{spatial id loss} we described finetuning Qwen2-2B with a spatial ID augmented loss module. Specifically, we freeze all weights except the MLPs of the last six vision encoder blocks, which we believe are most important for spatial reasoning, and train with synthetic data akin to those shown in Fig. \ref{fig:imgs for synth app}. 
We batch 15 images of the same object but varying locations into a mini-batch, and compute the predicted spatial ID by subtracting the average activation. This is similar to how we extracted the spatial IDs in \S\ref{sec: empirical_extraction}.

\begin{figure}[h]
    \centering
    \includegraphics[width=0.4\linewidth]{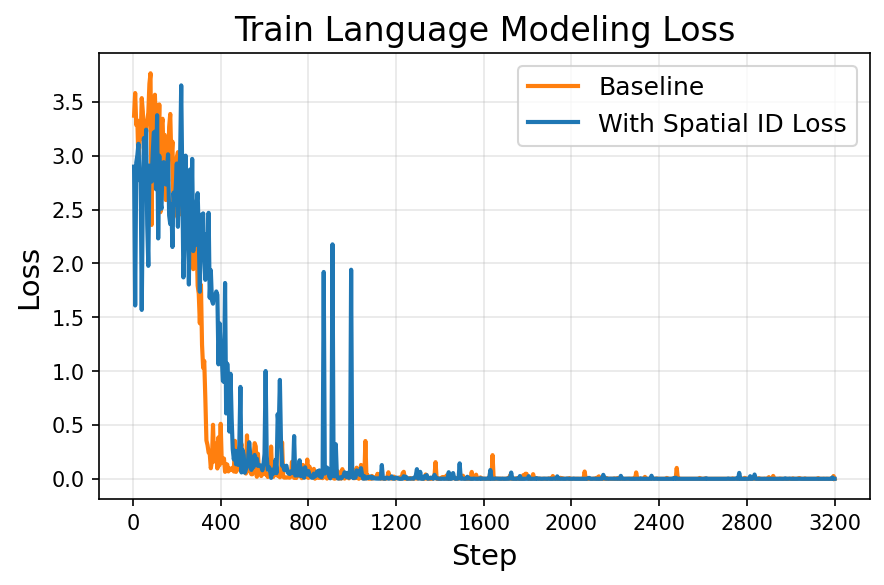}
\includegraphics[width=0.4\linewidth]{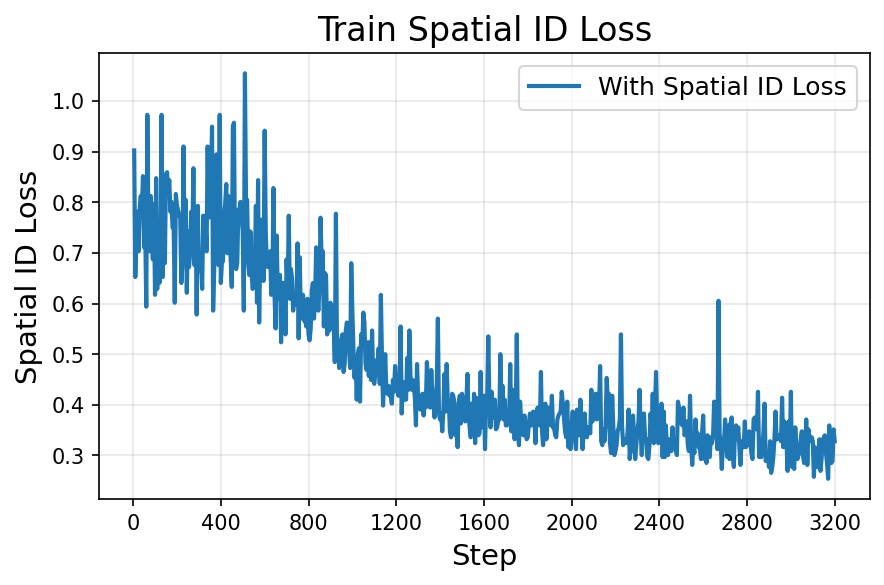}
\includegraphics[width=0.4\linewidth]{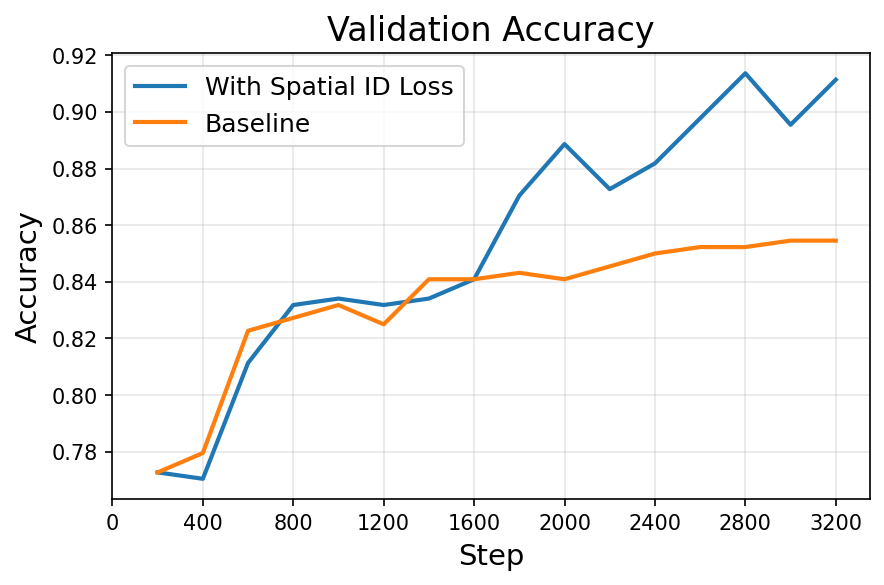}
    \caption{Plots from Qwen2-2B finetuning with and without spatial ID loss}
    \label{fig:qwen finetuning}
\end{figure}

The validation accuracy and training plots are shown in Fig. \ref{fig:qwen finetuning}. We see that with spatial ID loss, model accuracy on the naturalistic validation set (COCO-spatial) increases around 6\% (absolute) beyond the baseline plataeu, reaching a 90\% accuracy in under 2.8k steps.

\newpage

\section{Experimental Results on More Models}
\label{exp res on more models appendix sec}

\subsection{Spatial Grids on More Models}

\begin{figure}[H]
    \centering
    \includegraphics[width=1.0\linewidth]{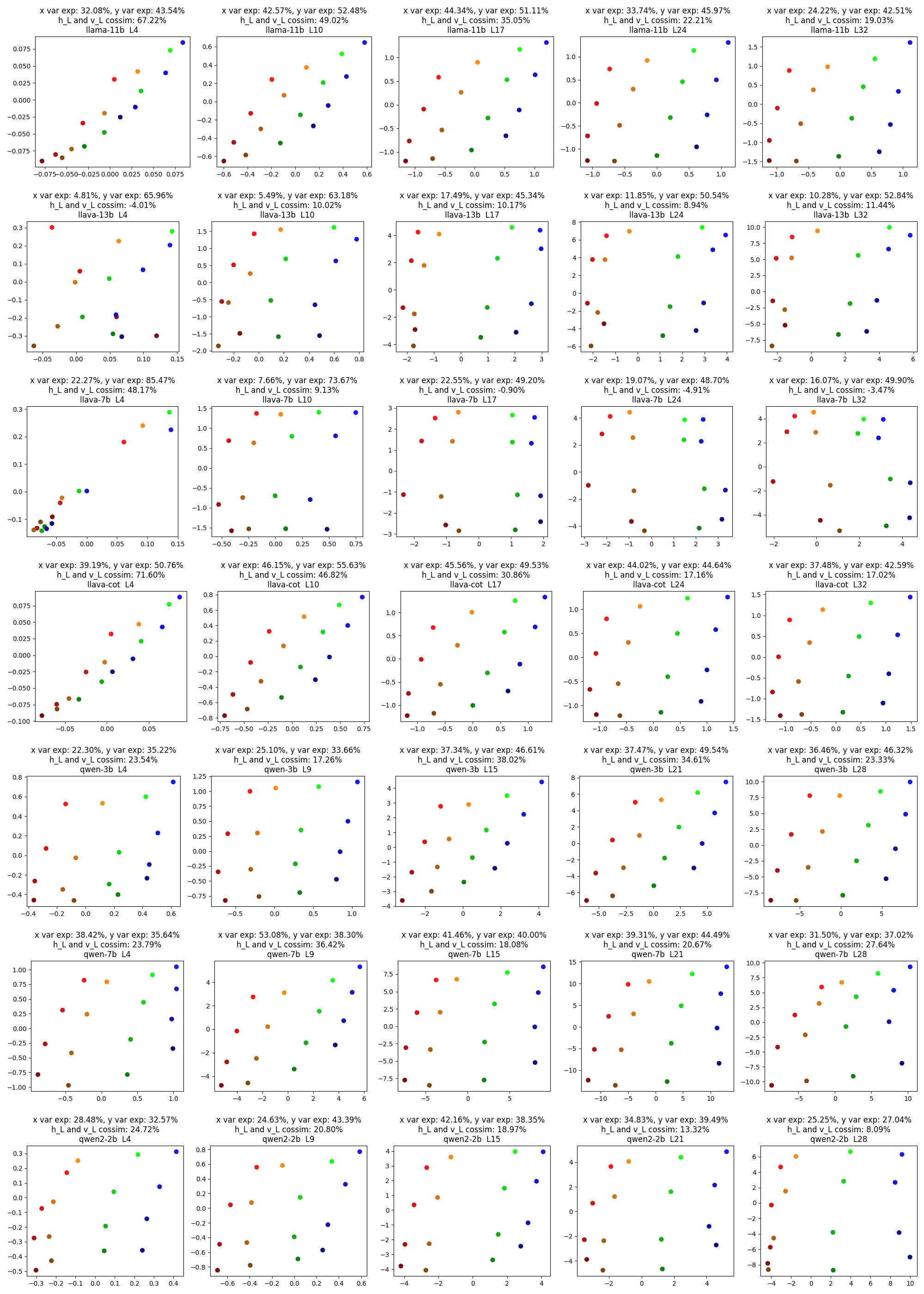}
    \caption{Spatial ID grids for LLaVA, LLaMA, and Qwen models.}
    \label{fig:spatialid grid supp 1}
\end{figure}

\begin{figure}[H]
    \centering
    \includegraphics[width=1.0\linewidth]{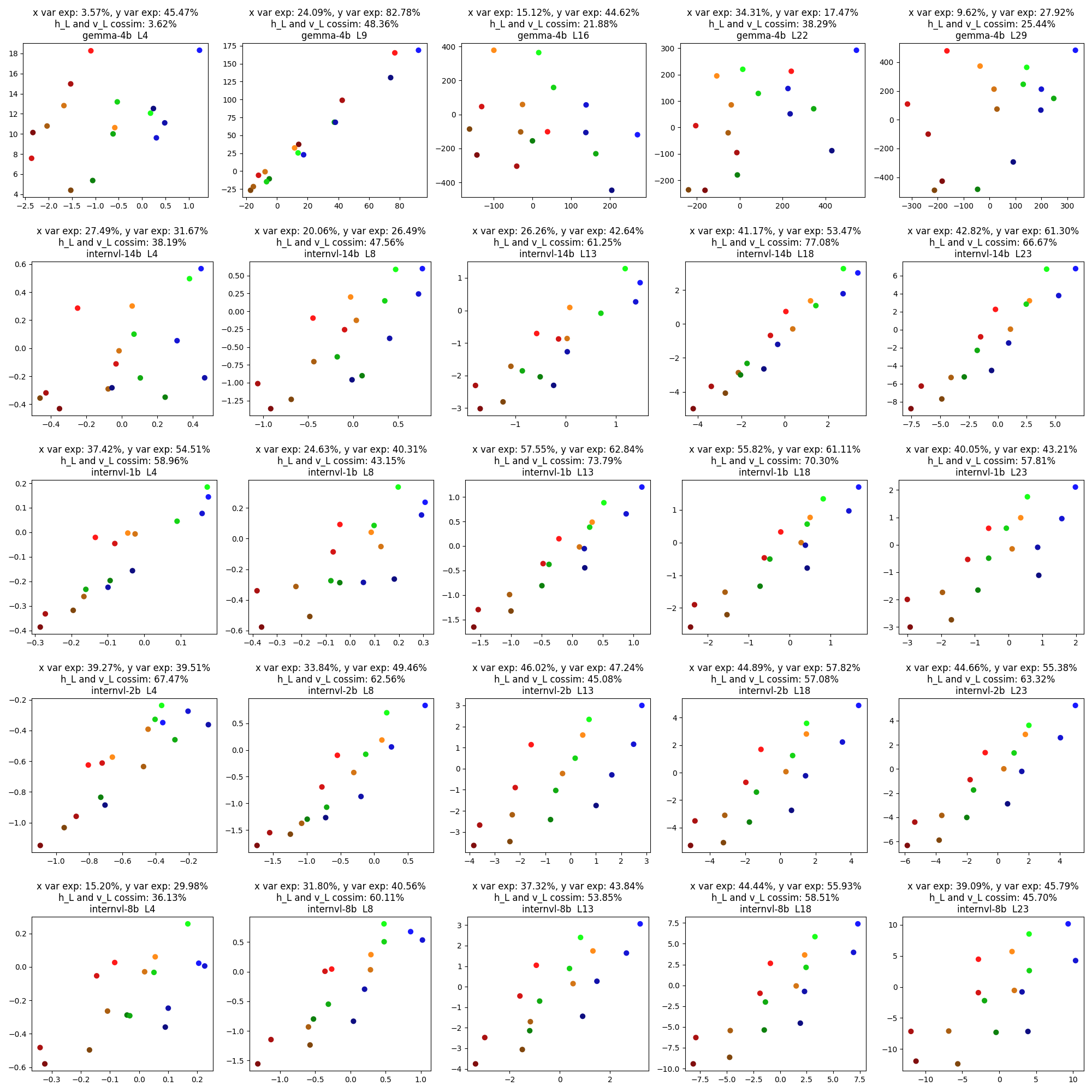}
    \caption{Spatial ID grids for Gemma and InternVL models.}
    \label{fig:spatialid grid supp 2}
\end{figure}

Fig. \ref{fig:spatialid grid supp 1},\ref{fig:spatialid grid supp 2} show spatial ID grids for all models shown in Fig. \ref{fig:all_models}. Subplot headings include \% variance explained by each spatial axis, as well as the cosine similarity between the spatial axes.
Notably, spatial IDs on some models seem to yield highly correlated $v_L$ and $h_L$, suggesting different spatial directions may be conflated.
\newpage

\subsection{Temporal Grids on more Models}
\label{supp sec temporal grids more models}

\begin{figure}[H]
    \centering
    \includegraphics[width=1.0\linewidth]{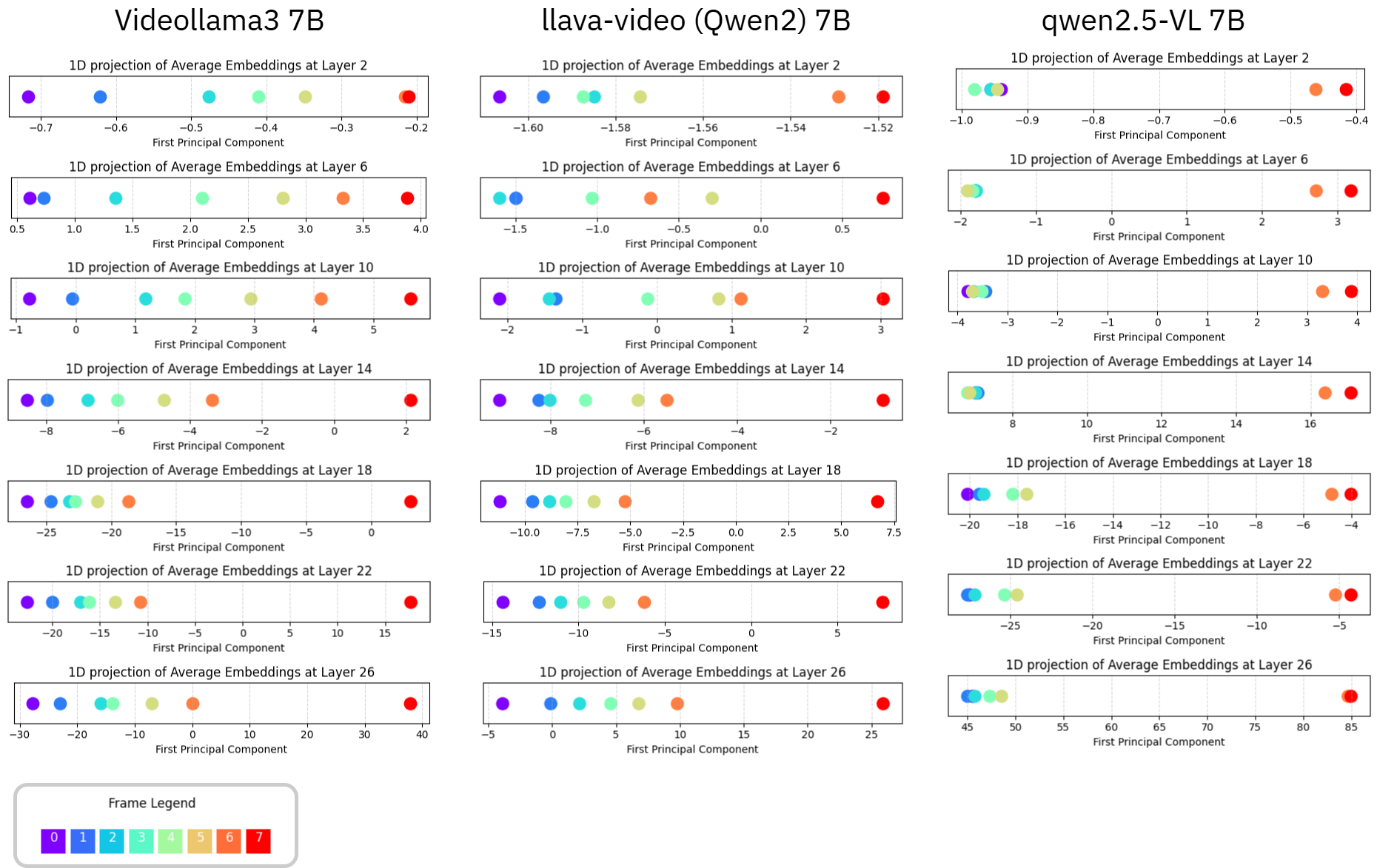}
    \caption{Temporal ID grids for \vidllama, \llavavid, and Qwen2.5. }
    \label{fig:video all ids}
\end{figure}

Across the models, there is a trend for the last frame(s) to be much farther away from the rest of the frames' temporal IDs. This may be a result of the data bias during model training, where a lot of instruction tuning datasets will ask temporal questions that only require paying attention to the last frame (e.g., \textit{did the person leave the room?} only requires looking at the first and last frame, and intermediate nuances are less important).

\section{Counterfactuals}
\label{supp sec counterfactuals}

\subsection{Spatial IDs from non-object words}

In \S \ref{sec information flow} we concluded that spatial information is largely stored in object words at intermediate layers. But could the information storage be spread out across the sequence dimension in internal activations?
To test this, we extract spatial IDs from non-object words, per \S \ref{sec: empirical_extraction}. Specifically we choose the spatial words in the query format "Is the \{obj\_word1\} \{spatial\_word1\} \{spatial\_word2\} \{obj\_word2\}?". 

We then perform steering on object words, as well as non-object words, with both the spatial IDs extracted from object words and non-object words. We use the same steering algorithm as Alg. \ref{alg:intervene_on_obj}. The results are shown in Table \ref{tab:counterfactual}. We see that some spatial semantics seems to be extractable from non-object words, and model belief is partially steerable through non-object words when using spatial IDs from object words, likely due to the fact that semantic word meanings are rarely perfectly contained within the initial word token in practical applications. In particular, spatial word tokens are likely to have information bleed over from the object word tokens while performing spatial queries, due to the way attention merges information between similar sequences. Regardless, effects from steering on object words with spatial IDs from object words is by far the strongest.

\begin{table}[H]
    \centering
    \begin{tabular}{c|c c c c}
    \toprule
    Model Name & ID-LR/apply-LR & ID-LR / apply-obj & ID-obj / apply-obj & ID-obj / apply-LR \\
    \midrule
         Qwen-3b & 18.77 & 51.19 & \textbf{81.23} & 48.46 \\
         Qwen-7b&  6.12 & 38.78 & \textbf{72.35} & 47.96 \\
         LLaVA-7b & 29.83 & 30.51 & \textbf{46.26} & 26.19 \\
         LLaVA-13b & 8.16 & 19.39 & \textbf{48.30} & 15.25\\
         \bottomrule
    \end{tabular}
    \caption{Spatial IDs extracted from object words and applied onto object words are most successful at steering model beliefs. Spurious effects are observed from IDs extracted from or applied unto unrelated words, but the effects are clearly concentrated on the object words. }
    \label{tab:counterfactual}
\end{table}

\subsection{Mirror Swapping on non-object words}
\label{mirror swapping on non obj}
\begin{figure}[h]
    \centering
    \includegraphics[width=0.89\linewidth]{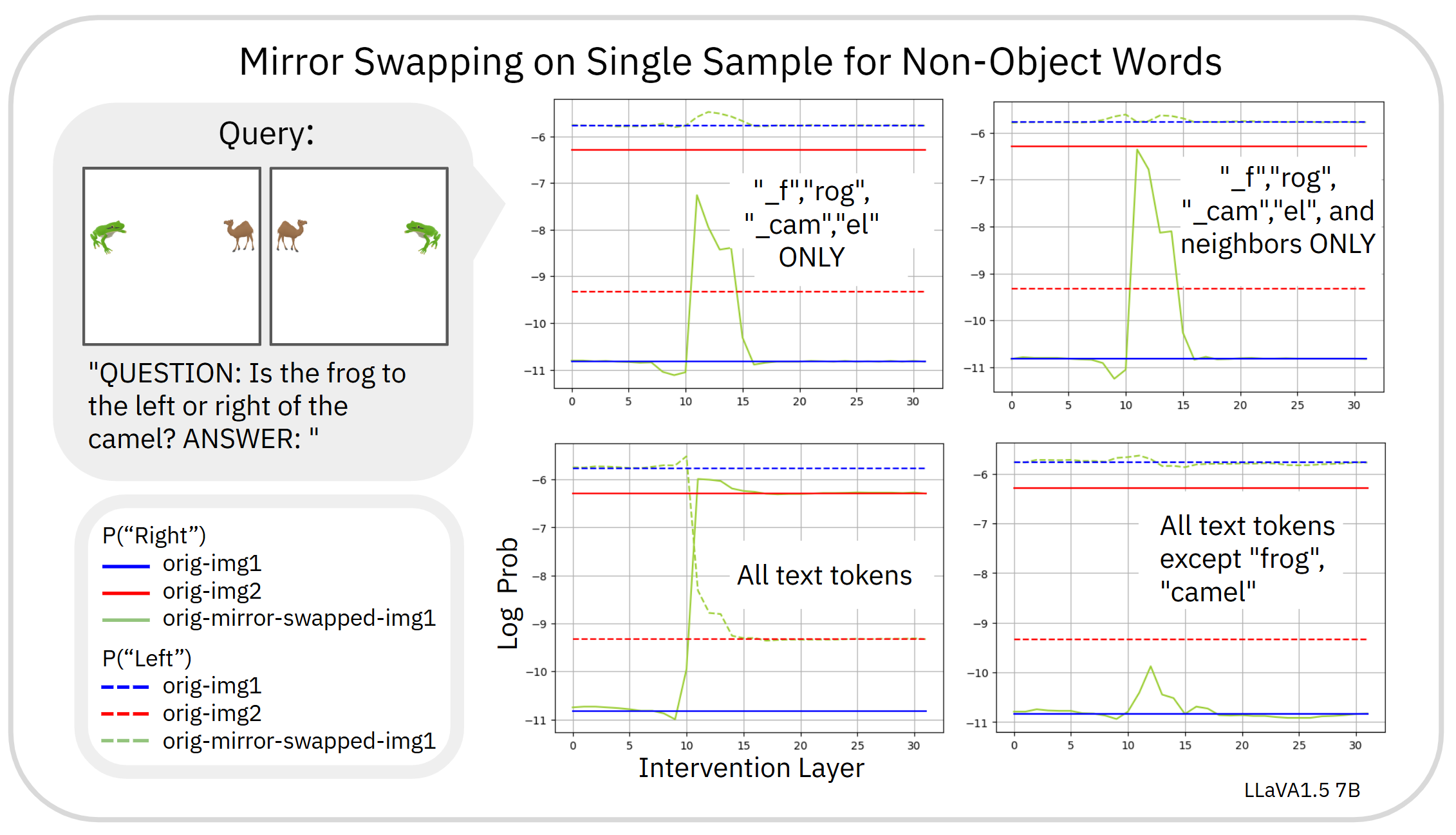}
    \caption{Synthetic image example for mirror swapping. Swapping non-object tokens has minimal impact on model belief change.}
    \label{fig:single sample control}
\end{figure}

To first showcase on a single sample the difference between mirror swapping on object tokens versus non object tokens, we choose a synthetic example with two objects on a blank background. 
 Fig. \ref{fig:single sample control} shows the results. 
 Here, the green line shows model belief change, and the x axis indicates the layer of intervention. Mirror swapping at just the object words has a slightly less prominent effect than intervening at the object words in addition to immediate neighboring tokens (such as the space preceding the animal word), which captures some of the information bleed. Swapping all tokens except for those belonging to object words, on the other hand, has the smallest observable effect. Hence spatial information is likely concentrated in object tokens.

\begin{figure}[h]
    \centering
\includegraphics[width=0.5\linewidth]{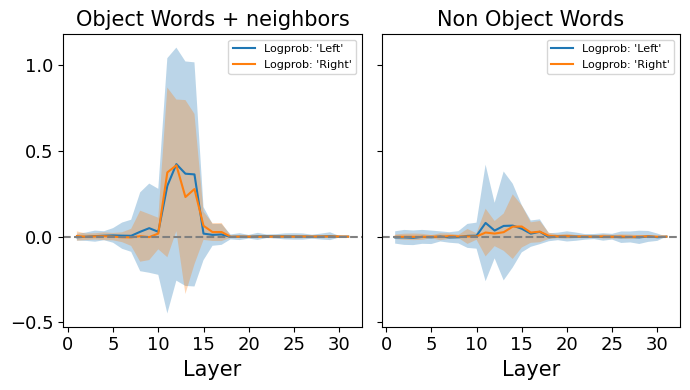}
    \caption{Mirror Swapping non object words}
    \label{fig:non obj words swap}
\end{figure}

We can now repeat the mirror swapping at non-object tokens at scale on COCO images. Fig. \ref{fig:non obj words swap} shows the difference between steering on object words and immediate neighboring tokens, versus non object words. Here, the non object words are randomly selected to be the same number of token indices as the object words. We again see that while there is some minor information bleed, the bulk of spatial ID information lies in object word tokens.

\subsection{Steering Effects on Orthogonal Directions (x vs. y), (time vs. x)}

\begin{figure}[H]
    \centering
    \includegraphics[width=0.5\linewidth]{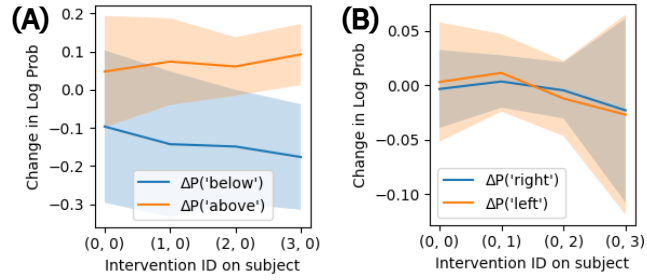}
  \caption{Steering effects of horizontal vectors on vertical beliefs (A) and vertical vectors on horizontal beliefs (B) in LLaVA.}
  \label{counterfactual_orthog}
\end{figure}

In Fig. \ref{fig:Steering-res}, we show the results of horizontal steering on ``left" vs. ``right" beliefs, and vertical steering on ``above" and ``right". To verify that steering directions can be decoupled, we perform the same steering and observe affects on beliefs of orthogonal directions. We show results of this preliminary analysis on LLaVA. Fig. \ref{counterfactual_orthog} shows these orthogonal effects. Spatial IDs that are equivalent in the y coordinate but changing in x coordinate do not change beliefs in ``above" or ``below". Similarly, static x coordinates with a changing y coordinate in spatial IDs has no effect on model belief about ``left" and ``right". 

\begin{figure}[H]
    \centering
    \includegraphics[width=0.4\linewidth]{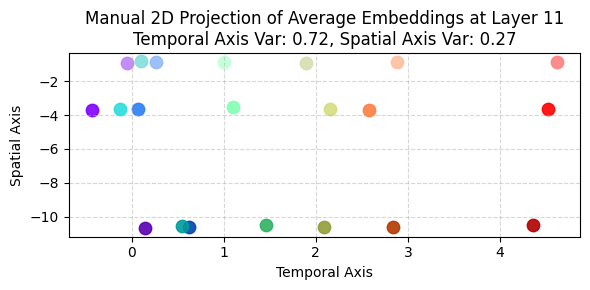}\includegraphics[width=0.4\linewidth]{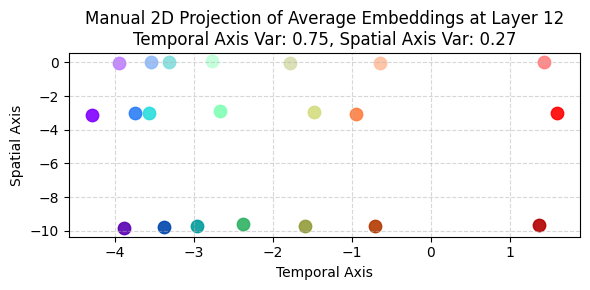}
    \\
\includegraphics[width=0.4\linewidth]{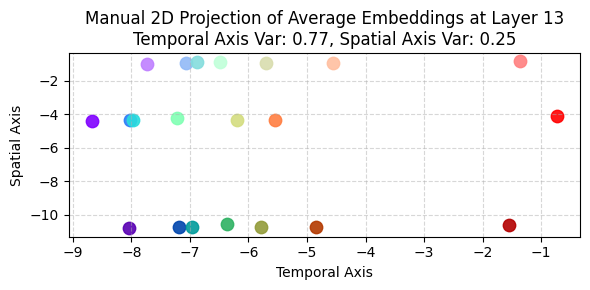}\includegraphics[width=0.4\linewidth]{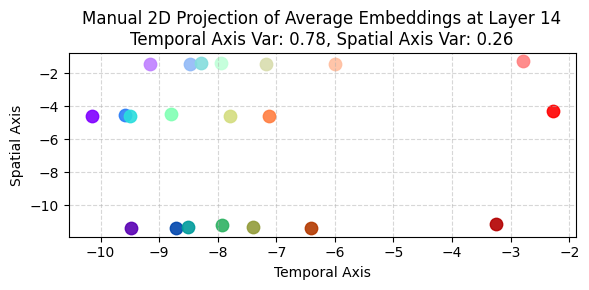}
    \caption{Spatiotemporal ID grid, where y axis is space and x axis is time.}
    \label{fig:video space time}
\end{figure}

Further, we check orthogonality between the space dimension and temporal dimension in video models. Fig. \ref{fig:video space time} shows a spatiotemporal ID grid from L11-14 on LLaVA-Video. The IDs are from videos where the object was in one of 8 frames (temporal change), and in one of 3 locations (spatial change). The experimental setup was minimal due to compute limitations. But even in this minimal setting, we see that the spatial and temporal axes are well separated.

\newpage

\section{Ablations}
\label{supp sec ablations}

\subsection{Scaling Analysis for Spatiotemporal ID Extraction}
\label{scaling supp sec}

\begin{figure}[H]
    \centering
    \includegraphics[width=1.0\linewidth]{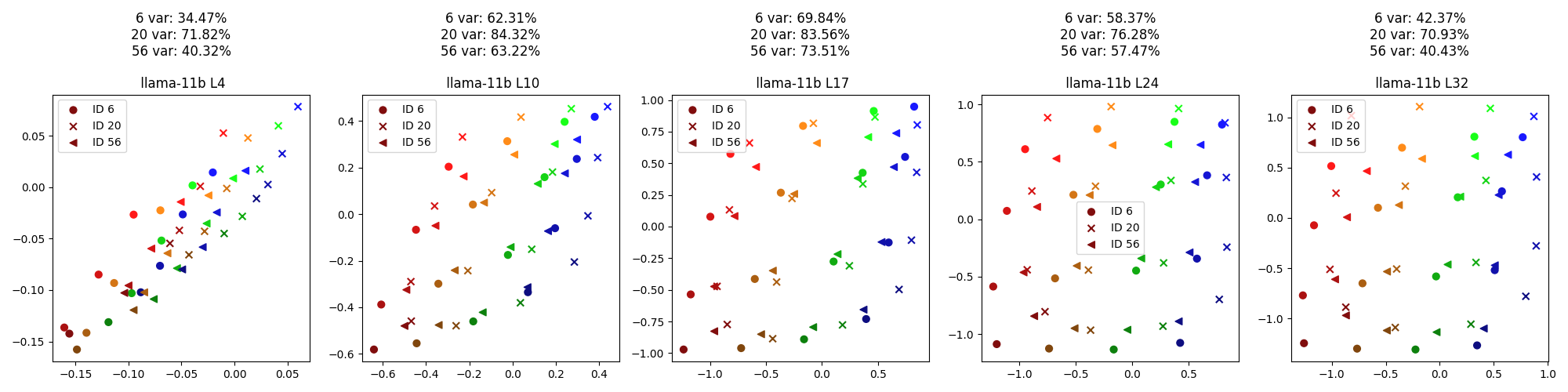}
    \includegraphics[width=1.0\linewidth]{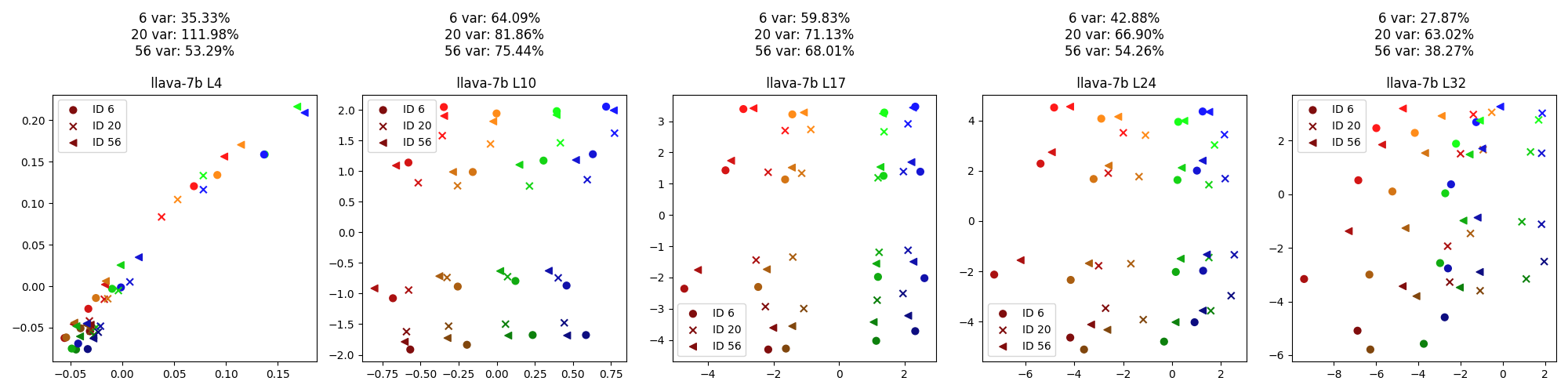}
    \includegraphics[width=1.0\linewidth]{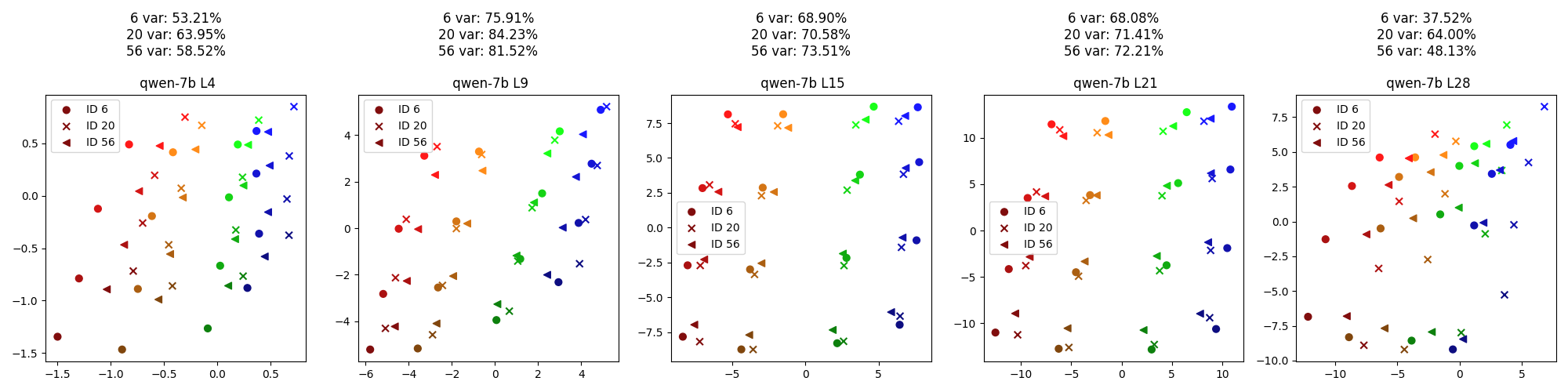}
    \caption{Extracting IDs with 6, 20, and 56 object pair images.}
    \label{fig:scaling analysis}
\end{figure}

The projection axes are from the 56 object pair case. At intermediate layers, where we expect spatial IDs to be most crucial, we see a tight color-wise clustering, indicating spatial IDs extracted from various numbers of objects still converge.
The variance explained by the spatial axes for all spatial ID extraction cases is $\gtrsim 50\% $, showing even at as little as 9 object pairs, we can extract good spatial IDs.

\subsection{Varying prompt wording and object sizes during extraction}

\begin{figure}[H]
    \centering
    \includegraphics[width=1.0\linewidth]{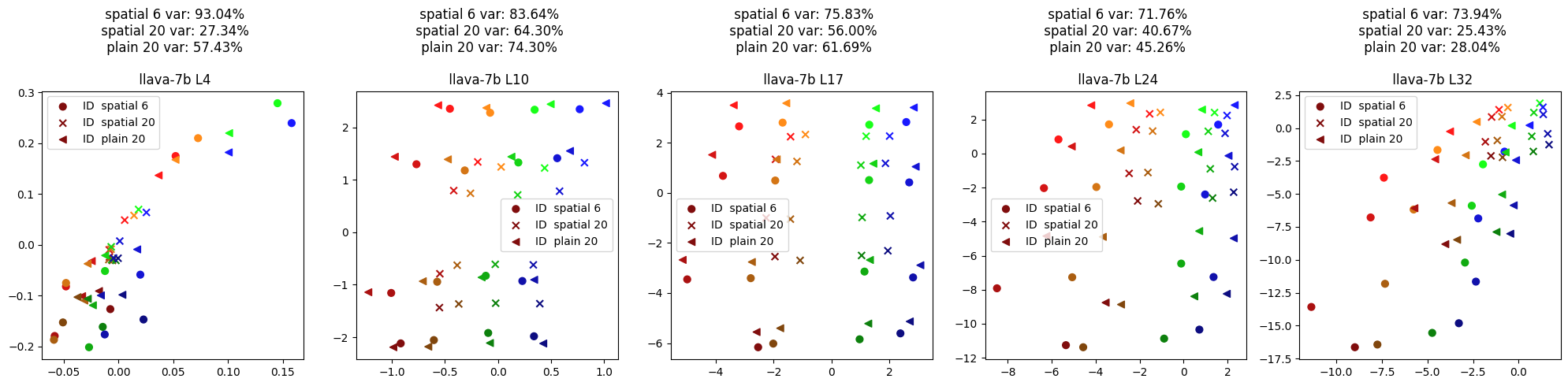}
    \caption{Plain prompts and spatial prompts projected onto spatial axes created from spatial prompts. Colors exhibit tight clustering.}
    \label{fig:plainprompt}
\end{figure}

\textbf{Varying prompt wording}. In this work, we use a spatial query in the form ``Is the x to the left or right of the y?" to extract spatial IDs from object words. To verify that the choice of prompt does not matter, and that information about spatial location of objects flows into the word activation regardless, we extract spatial IDs from a \textit{plain} prompt in the form ``Is there an x or y in the image?". In Fig. \ref{fig:plainprompt} we show the results of this extraction. We see that regardless of the input query formatting, spatial information can be extracted from the object words at intermediate.

\begin{figure}[H]
    \centering
    \includegraphics[width=0.7\linewidth]{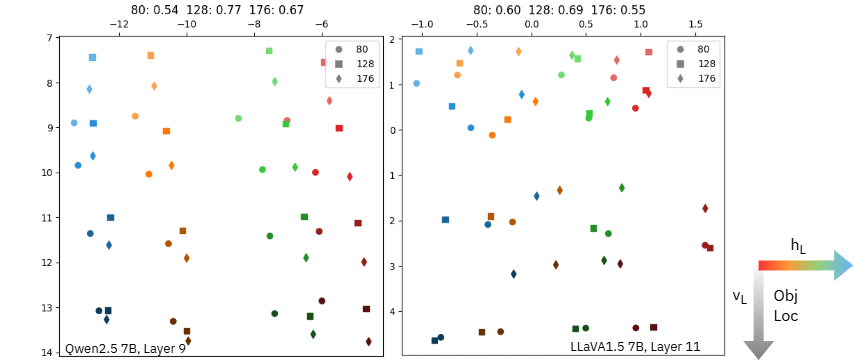}
    \caption{Spatial ID grids for Qwen and LLaVA, extracted from multiple object sizes. Circles are IDs extracted from images where object size was 80px in diameter, square is 128px, and diamond is 176px. On the top row, is the variance explained by the spatial axes for each size case.}
    \label{fig:supp_sizes}
\end{figure}

\textbf{Varying Image Sizes}. To test that spatial IDs are roughly agnostic to object size, we extract spatial IDs from images where the object is 80px in diameter, 128px, and 176px, then project all extracted spatial IDs onto spatial axes created only from the medium sized object case. The result is shown in Fig \ref{fig:supp_sizes}. While the variance explained by the spatial axes drops by 10$\sim$20\%, the sptial IDs extracted from different sized objects still exhibit strong in-color clustering, and $\gtrsim 50\% $ of variance are explained by the spatial axes. 

\section{Theoretical Analysis of Spatial IDs}

\subsection{Informal Proof for Spatial ID Emergence}
\label{supp sec informal_proof}

\textbf{Proposition:} \textit{Universal spatial IDs arises in any VLM using positional encoding, per self attention \citep{vaswani2017attention}.}

\textbf{Preliminaries}.
Consider a VLM layer with one attention head. Let the input sequence contain
projected visual tokens $\{x_p\}_{p\in\mathcal P}$, where each patch index
$p=(i,j)$ lies on an $m\times m$ grid, and text tokens including an object
token $o$ (as in prompts ``Is there an $o$?''). Define queries, keys, and values
$q_o=W_Q r_o$, $k_p=W_K x_p$, $v_p=W_V x_p$, and the standard residual update
$r_o \gets r_o + W_{out} \!\sum_{p}\alpha_{o\leftarrow p} v_p$ with
$\alpha_{o\leftarrow p}=\mathrm{softmax}_p(q_o^\top k_p/\sqrt d)$.

We make two very weak assumptions.

(1) First, we approximate that each patch vector decomposes as
\[
x_p \;=\; s_p \;+\; P\,\psi(p) \;+\; \varepsilon_p ,
\]
where $s_p$ encodes content (semantics), $\psi(p)\in\mathbb R^{d_\psi}$ is a
shared positional basis (e.g., learned 2D embeddings or RoPE-induced features),
$P$ maps positional features into model space, and $\varepsilon_p$ is some small deviation. In practice, explicit positional encoding is appended in autoregressive VLMs, so this assumption is explicitly true. In \S \ref{appendix sec empirical_spatial_id_posenc} we show empirically that positional encodings of VLMs linearly explain spatial IDs.

(2) We also assume that at a patch level, objectness is still encoded such that for images where a visual instance of the object word $o$ occurs at a
unique patch $p^*=(i,j)$, the attention kernel is peaked at $p^*$. In other words, 
$q_o^\top k_{p^*}\gg q_o^\top k_{p}$ for $p\neq p^*$, so that
$\alpha_{o\leftarrow p^*}\approx 1$. Again, this is almost always true in practice, as modality alignment is encouraged during training.


\paragraph{Proof.}
Write the value at patch $p$ using the decomposition:
\begin{equation}
\begin{aligned}
v_p \;=\; W_V x_p
\;=\; W_V s_p + W_V P\,\psi(p) + W_V \varepsilon_p
\end{aligned}
\end{equation}

Under assumption (2), the attention update to the object token is
\begin{equation}
    \delta r_o \;=\; W_{out} \sum_{p}\alpha_{o\leftarrow p} v_p
\;\approx\; W_{out}W_V x_{p^\star}
\end{equation}

Then we can rewrite Eq.\ref{eq:object-specific-extract} as:

\begin{equation}
\label{eq:solvestep1}
\begin{aligned}
        \Delta^{(o)}_L(p^\star) = r_{o,p^*} - \overline{r_{o,p}}
        \\
        = \Big(r_o + W_{out}W_{V}(s_{p\star} + P \psi(p^\star) + \epsilon_{p\star}\Big) - \overline{\Big(r_o + W_{out}W_{V}(s_{p} + P \psi(p) + \epsilon_{p}\Big)}
        \\
        =W_{out}W_{V}P\Big(s_{p^\star} - s_p + \psi(p^\star) - \psi(p) + \epsilon_p^\star - \epsilon_p\Big)
\end{aligned}
\end{equation}

Note that $s_{(o,p^\star)} = s_{(o,p)}$ for any $p$, for the first initial text embedding. Therefore, we can reduce Eq. \ref{eq:solvestep1} into:

\begin{equation}
\label{solve:final eq}
        \Delta^{(o)}_L(p^\star) = \Delta^{(o)}_L(i,j) \simeq W_{out}W_{V}P\Big(\psi(i,j) - \overline{\psi(p)}\Big)
\end{equation}

This expression is independent of $o$ except through the common matrix $W_{out}U$,
so averaging over objects leaves it unchanged. (In practice, we perform the averaging to reduce background noise.)

Notice that $W_{out}W_{V}P = M$ is fixed for some frozen network, and independent of location. Hence the centered attention update to the object token recovers a fixed linear
ID of a shared positional basis, i.e., a universal spatial ID. 
The implications of the emergence of these intermediate IDs is that a shared spatial vocabulary need only be aligned with their respective positional basis vectors to perform ``reasoning".
Let $z_o$ be the residual stream at the object token after the update, and let
$W_{\mathrm{vocab}}$ be the (approximately linear) readout to logits. Then
\begin{equation}
    \ell(\textsc{left})-\ell(\textsc{right})
\;\approx\; (w_{\textsc{left}}-w_{\textsc{right}})^\top \Delta_L(i,j)
\;\approx\; (w_{\textsc{left}}-w_{\textsc{right}})^\top M \big(\psi(i,j)-\mu_\psi\big)
\end{equation}

so if $(w_{\textsc{left}}-w_{\textsc{right}})^\top M$ aligns with the
$x$-coordinate component of $\psi$, the model correctly predicts spatial words.

\paragraph{Multi-head and multi-layer accumulation.}
For $H$ heads, $M=\sum_{h=1}^H W^{(h)}_{\text{out}}W^{(h)}_V P^{(h)}$; across layers, the contribution composes linearly in the residual stream. The “alignment band” in our experiments corresponds to layers where $||M||$ (or its projection onto the readout) is largest.

\subsection{Empirical Relationship between Positional Encoding and Spatial IDs}
\label{appendix sec empirical_spatial_id_posenc}

\begin{figure}[H]
    \centering
\includegraphics[width=0.6\textwidth]{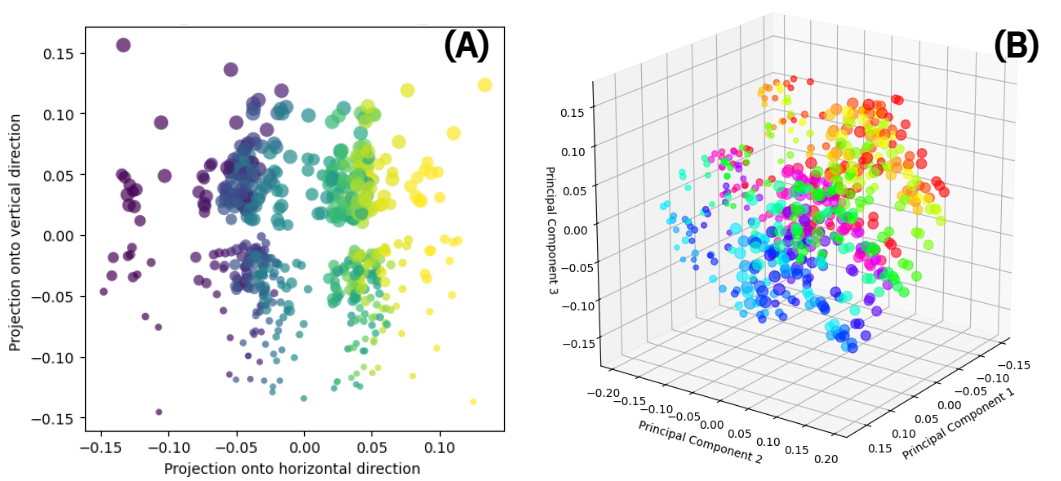}
  \caption{\llava Positional Encodings}
  \label{pos_enc_llava}
\end{figure}


Fig. \ref{pos_enc_llava} shows the patch level positional encodings from \llava (which uses the CLIP ViT-L/14 image encoder) projected onto 2 computed spatial axes or 3 principal components. The learned positional encoding vectors clearly have a linear structure, and with reduction in dimension are a linear transformation of the spatial ID grids we extract in \S \ref{sec: empirical_extraction}. For a model like Qwen, which starts with fixed Rotary Positional Encodings (RoPE) that are not learned, this separable structure is innate.
Previous work has shown that positional encoding in vision encoders continues to be linearly recoverable at penultimate activations 
\citep{ren2023masked}. 
We are interested in whether this structure is linearly recoverable in a downstream LLM, in the form of spatial IDs, to support \S \ref{supp sec informal_proof}. 
We show that for the models studied, there exist low rank linear mappings from positional encodings to spatial IDs.

\textbf{Setup}. 
Let $X \in \mathbb{R}^{Nxd}$ be positional encodings for some model and $Y \in \mathbb{R}^{NxM}$ be the spatial IDs extracted.
To find their linear relationship, we simply must solve for $W \in \mathbb{R}^{dxM}$ in $Y \approx XW$.

The least–squares solution is obtained with the Moore–Penrose pseudoinverse as 
$W^{\star} \;=\; X^{+} Y$.
To impose a rank constraint $r$, we compute the truncated singular value decomposition
$
X = U \Sigma V^\top
$ and
keep only the top $r$ singular values $\Sigma_r$.
Then the rank–$r$ solution is
\begin{equation}
    W_r = V_r \Sigma_r^{-1} U_r^\top Y
\end{equation}

The in–sample fit can be quantified by the coefficient of determination:
\begin{equation}
    R^2_r = 1 - \frac{\lVert Y - X W_r \rVert_F^2}{\lVert Y - \bar{Y} \rVert_F^2},
\end{equation}
where $\bar{Y}$ is the column–wise mean of $Y$.

For models like \llava and \llama, we acquire $X$ by taking the learned positional embeddings.
For models that use RoPE (which encodes position 
through complex rotations applied to query–key pairs) such as Qwen, we need an additional step to extract $X$.
Specifically, we can form a RoPE design matrix from the sinusoidal basis functions 
underlying these rotations and perform the same reduced–rank regression to the extracted spatial IDs $Y$.
Each position $p \in \{0,\dots,N-1\}$ is mapped
to sinusoidal features at different frequencies.  
Let the hidden dimension be $d$, with frequencies
\[
\theta_i = 10000^{-\tfrac{2i}{d}}, \quad i = 0,\dots,\tfrac{d}{2}-1.
\]

The RoPE design matrix $X_{RoPE} \in \mathbb{R}^{N \times d}$ is then
\begin{equation}
    X_{RoPE}(p) \;=\;
    \big[ \cos(\theta_0 p), \;\sin(\theta_0 p), \;
           \cos(\theta_1 p), \;\sin(\theta_1 p), \;\dots,\;
           \cos(\theta_{d/2-1} p), \;\sin(\theta_{d/2-1} p) \big],
\end{equation}
with each row of $\Phi$ corresponding to a position $p$.

\textbf{Results}.  We find that a weight matrix of rank 3 linearly relates the positional encoding matrix to the spatial IDs of a model with  $R^2 \geq 0.85 $. The three independent weight vectors likely correspond to horizontal, vertical, and radial axes, meaning such structure is preserved in the spatial IDs with high fidelity.
Results are shown in Table \ref{tab:lowranklinear}.

\begin{table}[h]
    \centering
    \begin{tabular}{c|c c}
    \toprule
        Model & Rank-2 $R^2$ & Rank-3 $R^2$  \\
        \hline
        
         LLaVA1.5-7B& 0.458 & 0.854
         \\
         LLaMA3.2VL-11B & 0.610 & 0.869
         \\
         Qwen2.5VL-7B & 0.605 & 0.903\\
         \bottomrule

         \hline
    \end{tabular}
    \caption{$R^2$ from low rank $W$}
    \label{tab:lowranklinear}
\end{table}

\section{LLM Usage Disclosure}
GPT-4 and GPT-5 were used in the process of occasionally coding experiments and editing paper wording.

\end{document}